%% file: main.tex
\documentclass[runningheads]{llncs}

\usepackage{accv}

\usepackage{accvabbrv}

\usepackage{graphicx}
\usepackage{booktabs}

\usepackage[accsupp]{axessibility}  % Improves PDF readability for those with disabilities.

\usepackage{siunitx}  % for \num command for rendering 5e-4 etc, correctly. 

\usepackage{hyperref}

\usepackage{orcidlink}

\usepackage{multirow}
\usepackage{placeins}
\usepackage{float}  % [H] placement
\usepackage{subcaption}
\usepackage{siunitx}
\usepackage{wrapfig}
\usepackage{tikz}
\usepackage{xcolor}

\newcommand{\tblsmall}[1]{{\footnotesize #1}}

\definecolor{myred}{HTML}{FFE4C4}
\definecolor{mygreen}{HTML}{EAFFEA}
\definecolor{myblue}{HTML}{E1E2FF}
\definecolor{mymagenta}{HTML}{FFE1FF}

\begin{document}

\title{Rethinking Uncertainty Quantification and Entanglement in Image Segmentation} 

\titlerunning{Rethinking UQ and Entanglement}

\author{Jakob Lønborg Christensen\inst{1}\orcidID{0009-0001-1510-6081} \and Vedrana Andersen Dahl\inst{1}\orcidID{0000-0001-6734-5570} \and Morten Rieger Hannemose\inst{1}\orcidID{0000-0002-9956-9226} \and Anders Bjorholm Dahl\inst{1}\orcidID{0000-0002-0068-8170} \and Christian F. Baumgartner\inst{2}\orcidID{0000-0002-3629-4384}}
\authorrunning{J.~Christensen et al.}

\institute{Technical University of Denmark, DTU Compute \and
University of Lucerne, Faculty of Health Sciences and Medicine, Switzerland}

\maketitle

\input{text/1intro}
\input{text/2background}
\input{text/3methods}

\input{text/4experiments}
\input{text/5discussion}

\clearpage  % Force all pending floats (figures/tables) to be placed before bibliography

\bibliographystyle{splncs04}
\bibliography{main}

\input{text/6supplement}

\end{document}

%% file: text/1intro.tex
% recommended to hit ~150 words (CURRENT 164 15:55 5 march)
\begin{abstract}
  Uncertainty quantification (UQ) is crucial in safety-critical applications such as medical image segmentation. Total uncertainty is typically decomposed into data-related aleatoric uncertainty (AU) and model-related epistemic uncertainty (EU). Many methods exist for modeling AU (such as Probabilistic UNet, Diffusion) and EU (such as ensembles, MC Dropout), but it is unclear how they interact when combined. 
  Additionally, recent work has revealed substantial entanglement between AU and EU, undermining the interpretability and practical usefulness of the decomposition. 
  We present a comprehensive empirical study covering a broad range of AU-EU model combinations, propose an entanglement proxy based on the relative performance of uncertainty measures, and evaluate model combinations across downstream uncertainty quantification tasks. Ensembles consistently show more favorable proxy values and superior performance. Softmax models usually beat other AU methods, except in calibration where the results are dataset-dependent. A softmax ensemble performs remarkably well on all tasks. Finally, we analyze potential sources of uncertainty entanglement and outline directions for mitigating this effect.
  \keywords{Image Segmentation \and Uncertainty Quantification \and Uncertainty Entanglement \and Aleatoric \and Epistemic}
\end{abstract}

\section{Introduction}
\label{sec:intro}

When using machine learning models in high-stakes environments such as medical imaging, knowing \textit{that} a model is uncertain is insufficient; we must also understand \textit{why}. Is the underlying data inherently ambiguous, or has the model simply not encountered enough similar examples?
Uncertainty quantification theory addresses this by distinguishing between aleatoric uncertainty (AU, or data-driven) and epistemic uncertainty (EU, or model-driven).

Separating uncertainty into aleatoric (AU) and epistemic (EU) components is useful because different downstream tasks are expected to depend on different sources of uncertainty. AU represents irreducible, data-inherent ambiguity, making it the natural signal for ambiguity modeling (AMB), where uncertainty estimates can be compared with disagreement among multiple annotators, as in Probabilistic UNet and SSN~\cite{prob_unet,ssn}. EU represents uncertainty due to limited knowledge from finite training data, making it relevant to out-of-distribution (OOD) detection, which distinguishes inputs unlike those seen during training~\cite{mc_dropout,deep_ensembles,ValUES}. Finally, calibration (CAL) should reflect all sources of predictive uncertainty, and therefore is expected to align with total uncertainty (TU), commonly additively defined with $TU=AU+EU$~\cite{kendall_gal}.

However, in practice, empirical AU and EU are often coupled, and the uncertainty measure that performs best for a task may not be the one theoretically expected to do so~\cite{uq_benchmark_classification,uq_entanglement_classification,ValUES}. Which measure performs best can depend strongly on the dataset and model choice~\cite{ValUES}. When the components are entangled, they no longer cleanly represent separate sources of uncertainty, obscuring what a model is uncertain about and potentially undermining downstream task performance and leading to safety-critical mistakes. Theoretical work has highlighted that entanglement may arise from the functional form of uncertainty measures~\cite{uq_measures} or from the loss minimization inherent to deep learning~\cite{uq_loss_pitfalls}. Moreover, large models can exhibit epistemic collapse, in which EU vanishes as network size increases~\cite{implicit_ensemble_collapse,uq_hole_collapse}, making TU increasingly correlated with AU. To advance safe machine learning, we therefore need to understand when and how this entanglement arises and develop strategies to mitigate it.

% Talk about necessity of modeling AU, EU or both.
AU is commonly modeled with probabilistic generative models~\cite{ssn,prob_unet,amb_diff_seg}, while EU is captured by inducing distributions over model parameters~\cite{swag,mc_dropout,deep_ensembles}. Although these can be combined~\cite{au_eu_clinical_trial,zepf_laplacian,au_eu_hyper_diffusion}, it remains unclear whether modeling both is necessary. Some studies suggest AU alone suffices~\cite{ValUES}, while others argue EU is important for OOD detection~\cite{zepf_laplacian}. Moreover, it is unclear which combinations of EU and AU approaches yield the best results. 

% our contributions
To address the many unknowns in the field, we provide a comprehensive empirical study focusing on the combination of aleatoric and epistemic uncertainty quantification for image segmentation. Our main contributions are as follows:
\begin{itemize}
    \item Building on the ValUES~\cite{ValUES} uncertainty quantification framework, we propose an entanglement proxy $\Delta$ based on whether the theoretically consistent uncertainty measure outperforms the alternative one for a given task.
    \item We evaluate $4 \times 5$ AU-EU model combinations on the LIDC-IDRI, MMIS NPC and Chákṣu IMAGE datasets across 3 downstream tasks.
    \item We provide task-specific model recommendations, balancing the entanglement proxy and performance.
    \item We analyze entanglement sources and suggest directions for mitigation.
\end{itemize}

%% TODO: intro figure

\begin{figure}[t]
    \centering
    \includegraphics[width=\textwidth]{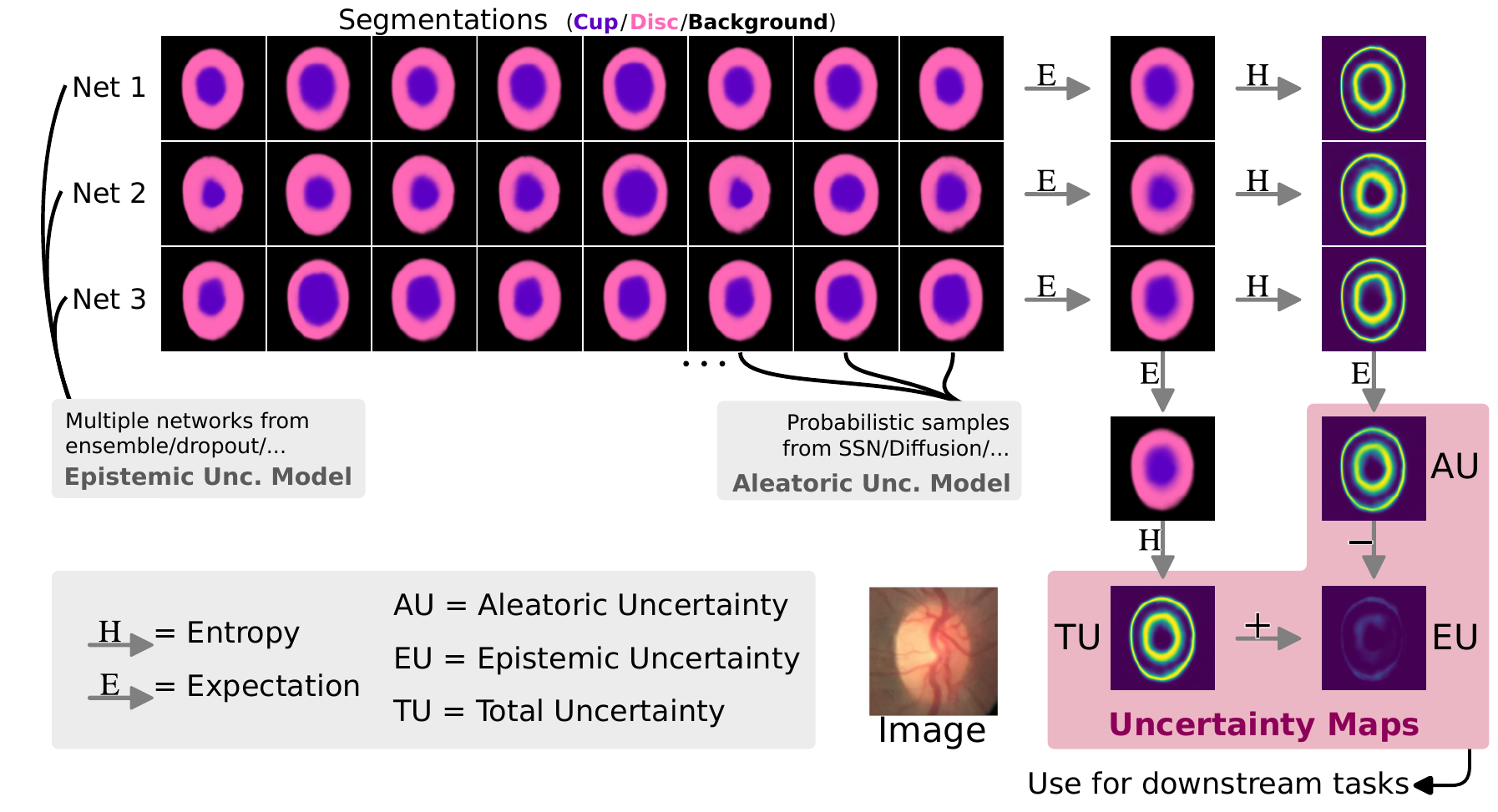}
    \caption{A visualization of aleatoric and epistemic modeling components and their aggregation into uncertainty measures for downstream tasks.
    }
    \label{fig:pred_grid_intro}
\end{figure}

%% file: text/2background.tex
\section{Background}
\label{sec: background}
% kendall & gal
Multiple approaches have been proposed to estimate aleatoric and epistemic uncertainty in deep learning~\cite{improved_it_measures,kendall_gal}. One influential approach is the information-theoretic framework proposed by Kendall and Gal~\cite{kendall_gal}, which models both types of uncertainty in a unified manner. This framework has been widely adopted in classification and image segmentation~\cite{zepf_laplacian,uq_benchmark_classification,uq_entanglement_classification}. Let $p=p(y|x,\theta)$ be the predictive distribution of a model for input $x$ and classification output $y$ given network parameters $\theta$. Then, the uncertainty components can be expressed as
\begin{equation}
\label{eq:uq_measures}
\begin{array}{c@{\hspace{1.2em}}c@{\hspace{1.2em}}c}
AU =
\underbrace{\mathbb{E}_\theta\!\left[ H\!\left(\mathbb{E}_y[p]\right) \right]}_{\text{\scriptsize Expected Entropy}},
&
TU =
\underbrace{H\!\left(\mathbb{E}_\theta\!\left[\mathbb{E}_y[p]\right]\right)}_{\text{\scriptsize Predictive Entropy}},
&
EU =
\underbrace{TU - AU}_{\text{\scriptsize Mutual Information}},
\end{array}
\end{equation}
where $H=\sum_i -p_i \log p_i$ is the Shannon entropy. We refer to $\mathbb{E}_y[p]$ as the \emph{generative mean}, the average predictive distribution over samples produced by a generative model for fixed parameters $\theta$. For deterministic models (``softmax''), the output itself is the generative mean, so the expectation over $y$ collapses. The equations are visualized in \cref{fig:pred_grid_intro}.

AU models for segmentation include the Stochastic Segmentation Network (SSN)~\cite{ssn}, Prob. UNet~\cite{prob_unet}, test-time augmentation~\cite{ayhan2018test} and diffusion models~\cite{seg_diff,amb_diff_seg}. These function independently of EU methods such as deep ensembles~\cite{deep_ensembles}, MC dropout~\cite{mc_dropout} and SWA/SWAG~\cite{swa,swag}, allowing straightforward combination~\cite{au_eu_clinical_trial,au_eu_hybrid_flow_depth_estimation}. For example, Zepf et al.~\cite{zepf_laplacian} combined SSNs with Laplacian networks, Wang et al.~\cite{au_eu_tta_ttd_medseg} combined test-time augmentation with dropout, and Chan et al.~\cite{au_eu_hyper_diffusion} combined diffusion with a Bayesian hyper-network. Which combinations synergize well for which tasks remains an open question.

% ValUES
Many works have demonstrated uncertainty entanglement~\cite{uq_entanglement_classification,uq_measures,uq_loss_pitfalls,uq_review}, with ValUES~\cite{ValUES} being the most comprehensive empirical study for image segmentation. ValUES evaluates a wide range of uncertainty quantification methods on multiple datasets and downstream tasks, showing that the theoretically consistent uncertainty measure is often not the one that performs best, with strong dataset and model dependency. While methods can model simulated toy data accurately, entanglement is prevalent on real data. ValUES also used variational model outputs (e.g.\ SSN) directly as the generative mean, arguing that AU and EU labels should be swapped since sample variability captures AU rather than EU. A major limitation of ValUES is that it neglects the interplay between AU and EU by using either aleatoric (SSN or test-time augmentation) or epistemic (dropout or ensemble) approaches separately rather than in combination. Furthermore, ValUES simulated aleatoric uncertainty via random label flipping on Cityscapes~\cite{cityscapes}/GTA V~\cite{gta}, which is a poor simulation since a simple hand-coded model head could capture this ambiguity perfectly.

%% file: text/3methods.tex
\section{Methods}
\subsection{Data}
Uncertainty quantification methods for segmentation can be optimally evaluated using data with multiple annotations per image. Such data captures aleatoric uncertainty through annotator disagreement, and therefore allows for more principled evaluation of AMB and CAL~\cite{prob_unet}. In our benchmark, we use the three datasets listed in \cref{tab:datasets}, all of which are resized to a $128\times128$ resolution. 

\begin{table}[t]
    \centering%training set
    \begin{tabular*}{\textwidth}{@{\extracolsep{\fill}}lccccccc}
    \toprule
    Dataset & Modality & $n_c$ & \#Im & Ann/Im & Train & Val & Test (ID/OOD) \\
    \midrule
    LIDC-IDRI & CT & 2 & 15096 & 4 & 9355 & 2689 & 3052 / $3052 \times 3$ \\
    MMIS NPC & MRI & 2 & 2260 & 4 & 1483 & 302 & 475 / $475 \times 3$ \\
    Chákṣu IMAGE & Fundus imaging & 3 & 1345 & 5 & 648 & 162 & 264 / 271 \\
    \bottomrule
    \end{tabular*}
    \caption{Summary of the datasets used in our experiments ($n_c:=$number of classes).}
    \label{tab:datasets}
\end{table}

\begin{figure}[t]
    \centering
    \includegraphics[width=\textwidth]{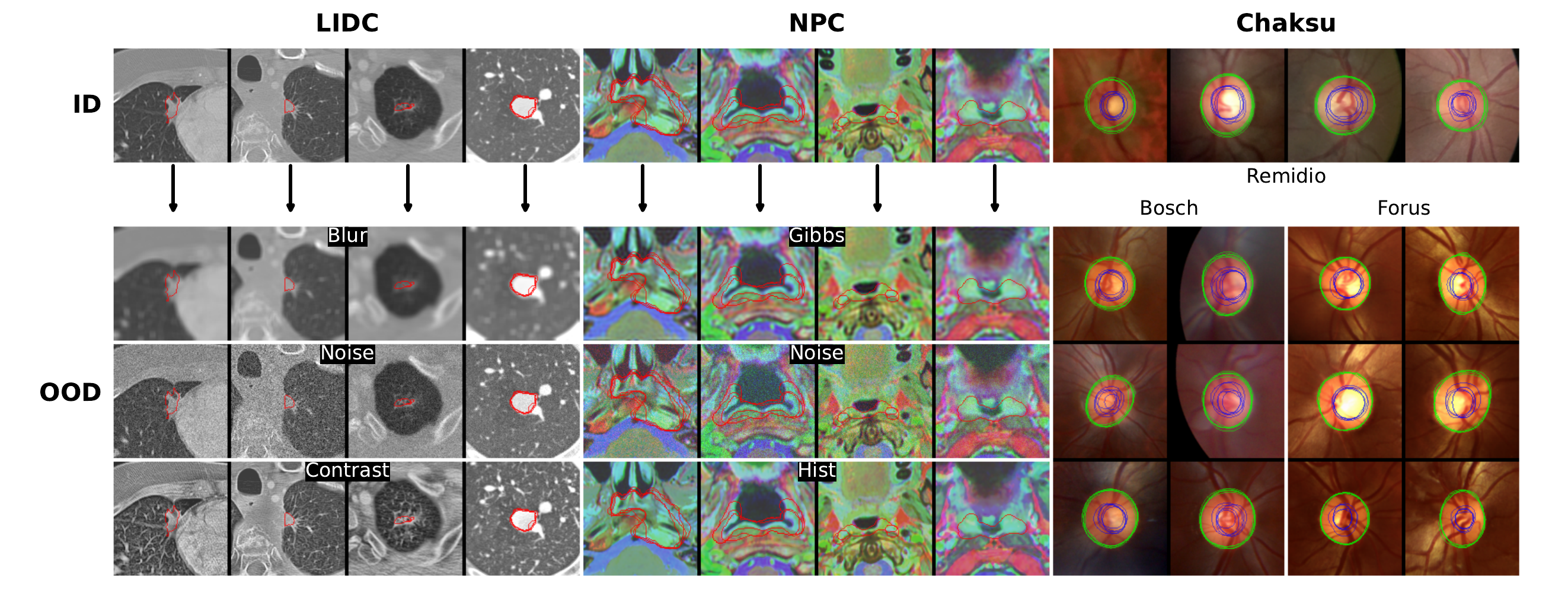}
    \caption{Example images from the three datasets, showing both ID and OOD images. The four binary-mask delineations are shown on LIDC and NPC as red lines, while the five cup and disc segmentations are shown as blue and green lines on Chaksu.
    }
    \label{fig:data_vis}
\end{figure}

The LIDC-IDRI dataset~\cite{lidc} (\textbf{LIDC}) consists of 1018 thoracic CT scans with annotated lung nodules (see \cref{fig:data_vis}), each annotated by at least four radiologists. For 2D segmentation, the 15,096 slices with exactly 4 annotations are commonly used~\cite{prob_unet,ssn,amb_diff_seg,amb_diff_seg_jakob}. Slices are lesion-centered crops, split per-patient into 60\%/20\%/20\% train/val/test sets. Annotator disagreement is dominated by nodule presence/absence rather than boundary delineation, so up to 3 segmentations can be empty. We apply three synthetic OOD shifts (blur, noise, contrast) to the test images, yielding one ID and three OOD test sets. 

The MMIS NPC-170~\cite{mmis} (\textbf{NPC}) dataset contains 170 multimodal magnetic resonance imaging (MRI) scans, 120 of which we can access. Each scan has aligned T1, T1-contrast and T2 sequences (shown as RGB in \cref{fig:data_vis}). Four senior radiologists independently annotated nasopharyngeal carcinoma gross tumor volumes for radiotherapy planning. The scans yield 2260 2D slices with at least one non-empty segmentation. Images are cropped square along the longer side and split per-patient into 64\%/16\%/20\% train/val/test sets. As OOD shifts, we apply three synthetic augmentations to the test set (Gibbs, noise, hist), emulating common MRI artifacts (Gibbs and Rician noise), while random histogram shifts resemble differences across scanning procedures and machines. Each channel is augmented separately because the channels stem from different scans.

The Chákṣu IMAGE dataset~\cite{chaksu} (\textbf{Chaksu}) consists of 1345 RGB fundus images with optic disc and cup segmentations (see \cref{fig:data_vis}), each annotated by 5 ophthalmologists. We keep only the largest connected component of each annotation to remove small artifacts. The data was captured with three devices: Remidio, Bosch and Forus (1074, 145 and 126 images, respectively). Remidio serves as ID and Bosch/Forus as OOD. We use the original test split (264 ID images) and split the remaining 810 ID images 80/20 into training and validation. 
Images are cropped around the optic disc and cup, with the crop side length set to twice the per-device average bounding box sidelength, ensuring the OOD-shift reflects image characteristics rather than size differences. 

\subsection{Aleatoric and Epistemic Uncertainty Models}
Here, we describe all models and hyper-parameter settings used in our experiments. 
We evaluate the following methods for AU estimation:
\begin{itemize}
    \item \textbf{Softmax}: A deterministic UNet trained with a standard cross-entropy loss. Referred to as "softmax" for the softmax output layer giving rise to the probabilities used in the uncertainty measures. 
    \item \textbf{Stochastic Segmentation Networks (SSN)}~\cite{ssn}: A stochastic segmentation network, modeling the logits as a Gaussian distribution with a mean and variance output by the network. We use a rank of 10 for the covariance matrix, $\epsilon = \num{1e-5}$ for numerical stability and 10 pretraining epochs estimating the mean before also modeling the variance.
    \item \textbf{Probabilistic UNet (Prob.\ UNet)}~\cite{prob_unet}: A probabilistic UNet with prior and posterior encoder networks, trained with KL divergence as an additional loss term. We use a latent dimension of 6, $\beta=\num{5e-4}$ (LIDC) and $\beta=\num{2.5e-3}$ (Chaksu) for the KL weight, with linear warmup over the first 32 epochs. Our $\beta$ values are smaller than typical because we use mean rather than sum reduction for the reconstruction term.
    \item \textbf{Diffusion}~\cite{DDPM,seg_diff,amb_diff_seg_jakob}: A diffusion model learning to reverse a noising process. We train with a uniformly weighted MSE loss in the data domain with a softmax head. The cosine noise schedule is used with input scaling~\cite{input_scaling} at $b=0.1$. For inference, we use the DDIM~\cite{DDIM} sampler with 10 timesteps.
\end{itemize}
\noindent
For the EU estimation we studied the following approaches:
\begin{itemize}
    \item \textbf{Deep Ensemble (Ensemble)}~\cite{deep_ensembles}: A deep ensemble of 5 models, each trained from different random initializations.
    \item \textbf{MC Dropout (Dropout)}~\cite{mc_dropout}: A model with channel-wise~\cite{spatial_dropout} dropout (p=0.2) before convolutional layers in residual blocks, at train and test time. 
    \item \textbf{Stochastic Weight Averaging-Gaussian (SWAG)}~\cite{swag}: SWAG approximates the weight posterior by fitting a Gaussian to weights collected during training. We use 30 snapshots from the last 30 epochs with AdamW instead of SGD, which performed better in early testing. The diagonal-only variant is referred to as \textbf{SWAG-D}.  
    \item \textbf{No EU}: No epistemic modeling; the AU model's generative outputs are used directly following ValUES~\cite{ValUES}, without swapping AU and EU labels.
\end{itemize}
\noindent
All model combinations are valid except softmax without EU (incompatible as it is deterministic). For each model, we adopt hyperparameters as specified in the original literature whenever possible. When hyperparameter values are ambiguous or may depend on the dataset or network architecture, we perform limited small-scale sweeps to select appropriate settings.

\subsection{Model Architecture}
To reduce architecture-based variation all models share a common attention UNet backbone based on Simple Diffusion~\cite{simple_diff,simple_diff2}. The probabilistic models require modifications to the base architecture. SSN adds lightweight heads for the diagonal variance and low-rank covariances of the Gaussian logits. Diffusion models add a timestep embedding. For Prob.\ UNet, two extra UNet encoders serve as the prior and posterior networks. To account for the added parameters, we scale the base channels for all three Prob.\ UNet networks from 32 to 24. The final model sizes are 15.78, 15.80, 15.81 and 16.89 million parameters for Softmax, SSN, Diffusion and Prob.\ UNet, respectively.

We train all models from scratch, as using pretrained weights would be difficult for the more involved AU models (Diffusion, Prob.\ UNet), complicating fair comparison. Models with No EU or SWAG/SWAG-D can share a single training run since they do not affect training directly.

\subsection{Uncertainty Quantification}

We generally follow the ValUES~\cite{ValUES} framework, basing experiments on the AMB, OOD detection and CAL tasks. At test time, we draw 10 AU samples and use 10 EU model instances, yielding up to 100 joint predictions per image for computing uncertainty maps via~\cref{eq:uq_measures}.

\textbf{Out-of-distribution detection.} Out-of-distribution detection separates in-distribution (ID) from out-of-distribution (OOD) samples using estimated uncertainty maps, and is measured by the area under the receiver operating characteristic curve (AUROC). An aggregation strategy is needed to reduce pixel-wise maps to image-level scores. ValUES suggested three strategies: the image-wise mean, the patch-level maximum uncertainty for a $10 \times 10$ patch, and a threshold-based strategy. Larger segmentations tend to have higher uncertainty under e.g. image-wise mean, and this can lead the model into predicting that all large lung nodules in LIDC are OOD and all small nodules are ID, which is undesirable. Therefore, we add the area-normalized and border-normalized strategies. The border length is the count of 1-connectivity neighboring pixel pairs with differing labels. If area or border is zero, we normalize by 1.

\textbf{Ambiguity modeling (AMB).} AMB aims to capture data-inherent variance. To measure how well this variance is captured, we compute pixel-wise variance over multiple ground truth annotations and compare to predicted uncertainty maps using normalized cross-correlation (NCC).

\textbf{Calibration (CAL).} The goal of calibration is to ensure that the predicted probabilities reflect the true likelihood of correctness. We use the expected calibration error (ECE), which is the average absolute difference between confidence and accuracy across 20 bins from 0 to 1, weighted by the number of pixels in each bin. We do not space the bins uniformly from 0 to 1 as is common, but instead use the (0\%, 5\%, 10\%\dots100\%) confidence quantiles as bin edges, avoiding situations where models can trivialize the score when all confidences are in a single bin. Uncertainty measures are converted to confidences by Platt scaling~\cite{platt_scaling} their negatives, with separate parameters for AU, EU and TU. Unlike ValUES, which averaged per-image Platt parameters, we compute dataset-wide parameters, fitted on the validation split and applied to the test split.

\begin{figure}[t]
%\begin{wrapfigure}{r}{0.33\textwidth}
    \centering
    \includegraphics[width=0.8\textwidth]{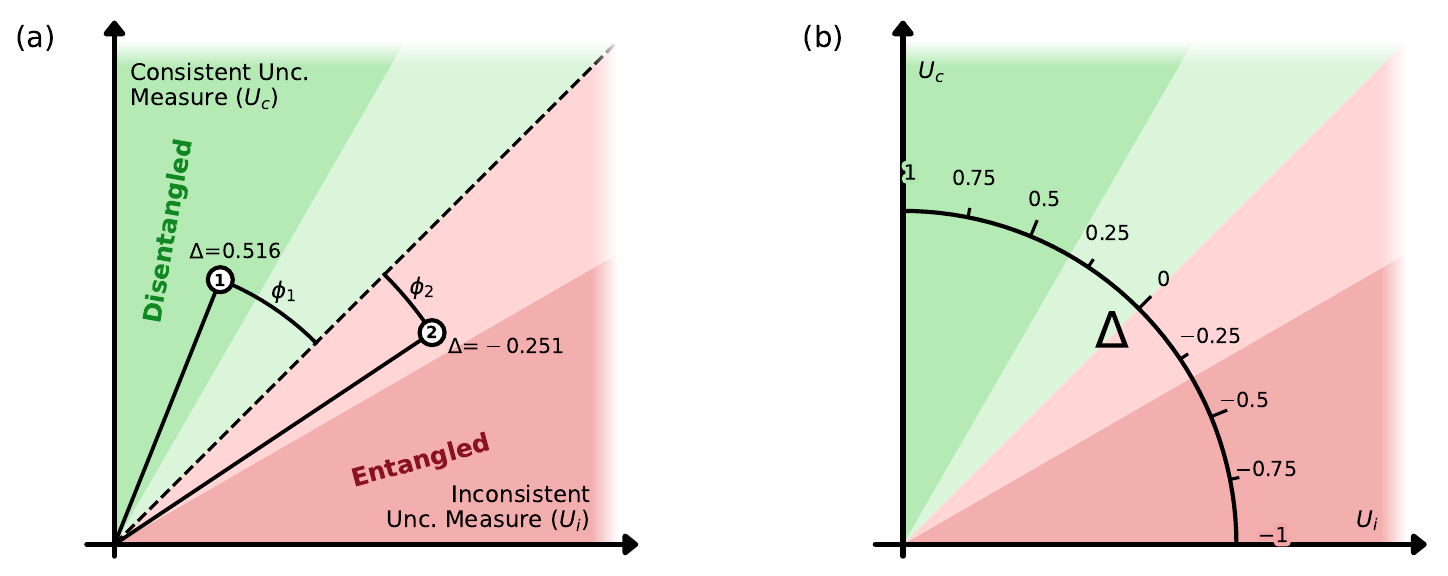}
    \caption{A visualization of the entanglement proxy $\Delta$. \textbf{(a)} Point 1 has a more favorable $\Delta$ value than point 2. The proxy is proportional to the signed angles ($\phi$). \textbf{(b)} Angles are converted to $\Delta$ based on the shown number line.}
    \label{fig:entangle_metric_vis}
%\end{wrapfigure}
\end{figure}

\textbf{Entanglement proxy ($\Delta$).} The empirical entanglement can be assessed through the relative performance of uncertainty measures. Specifically, if a theoretically consistent measure outperforms the theoretically inconsistent measure for a given task, this suggests coupling between them. To summarize this behavior, we define the entanglement proxy as:
\begin{equation}
    \Delta = s \frac{\arctan(U_c/U_i)-\pi/4}{\pi/4} = s \frac{\phi}{\pi/4},
\end{equation}
where $U_c$ and $U_i$ are the performance of the consistent and inconsistent uncertainty measures, respectively. For metrics where a lower value is better, such as ECE, we set $s=-1$ and otherwise $s=1$. The proxy ranges from $-1$ to 1, with larger values indicating better disentanglement (see \cref{fig:entangle_metric_vis}). It is worth noting that $\Delta$ does not directly measure statistical dependence between AU and EU and may be affected by the decomposition itself (\ie \cref{eq:uq_measures}), model capacity, or task characteristics, rather than entanglement alone. Therefore, absolute values should be interpreted with caution. However, the data, task, decomposition, evaluation protocol, and shared backbone are fixed in our comparisons, while the AU-EU model combination varies. Therefore, relative differences in $\Delta$ nevertheless provide a useful comparison of model combinations.
% maybe move to discussion?

\subsection{Training Details}

All models were trained with the AdamW optimizer, a learning rate of 1e-4, no weight decay, a batch size of 32 and gradient clipping with a maximum norm of 0.5. Models were trained for 1000, 320, and 500 epochs on LIDC, NPC, and Chaksu, respectively, with the exception of Prob.\ UNet models on LIDC, which were only trained for 500 epochs due to faster overfitting. Exponential moving average (EMA) checkpoints with a decay rate of 0.999 were used for evaluation. %On LIDC, the models were trained for 1000 epochs, except Prob.\ UNet models which were only trained for 500 epochs due to faster overfitting. All models trained on NPC and Chaksu were trained for 320 epochs and 500 epochs respectively. 

%% file: text/4experiments.tex
\FloatBarrier
\section{Results}
\begin{figure}[t]
    \centering
    \includegraphics[width=\textwidth]{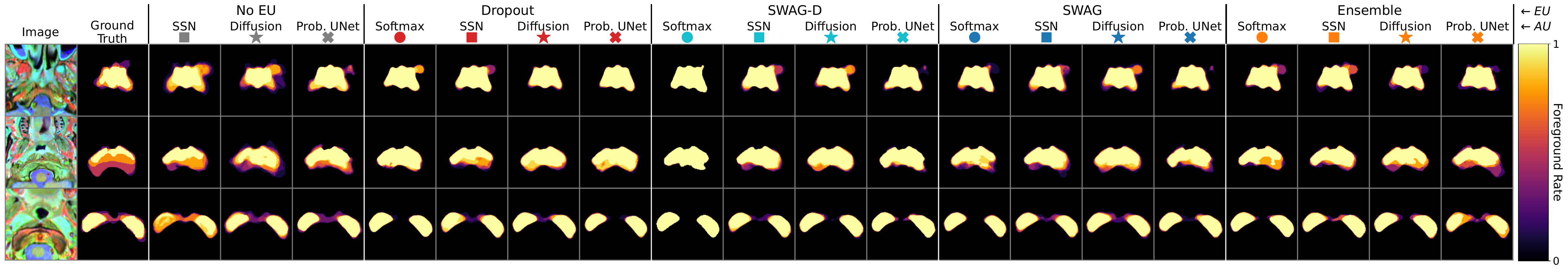}
    \caption{Mean model predictions ($\mathbb{E}_\theta[\mathbb{E}_y[p]]$) on NPC data (ID).}
    \label{fig:pred_grid_id_npc}
\end{figure}

\begin{figure}[t]
    \centering
    \includegraphics[width=\textwidth]{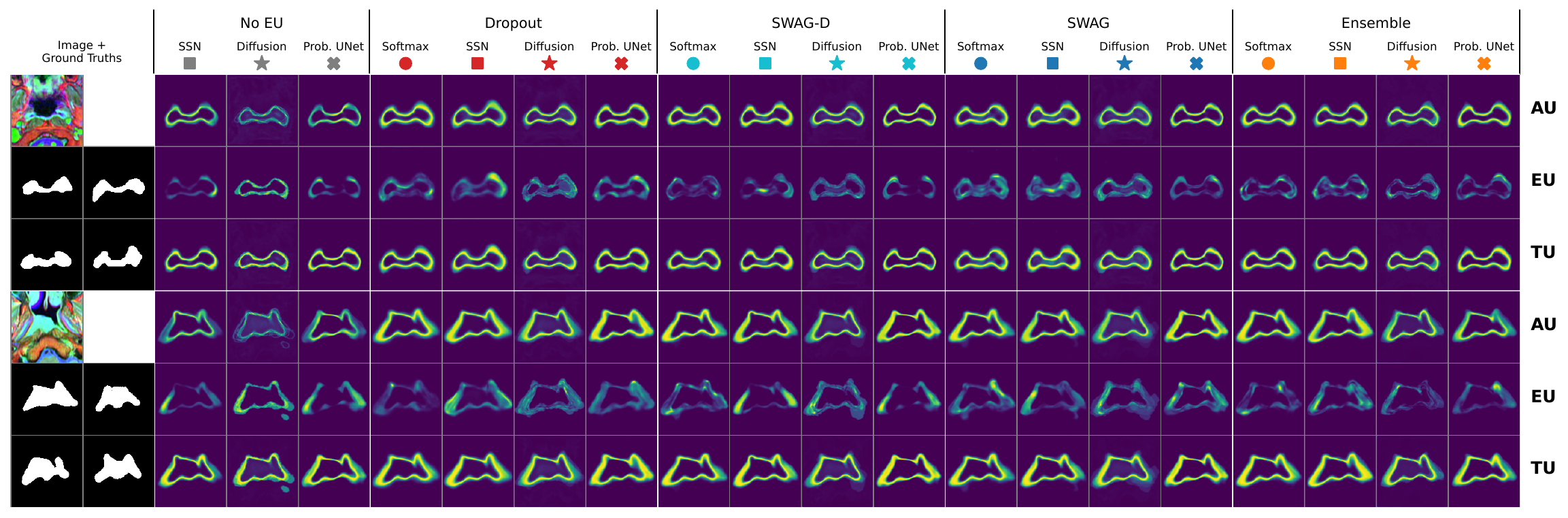}
    \caption{Aleatoric (AU), epistemic (EU) and total uncertainty (TU) maps for ID NPC images, with red numbers indicating the maximum value. 
    }
    \label{fig:unc_grid_id_npc}
\end{figure}

\begin{figure}[!htbp]
    \centering
    \includegraphics[width=\textwidth]{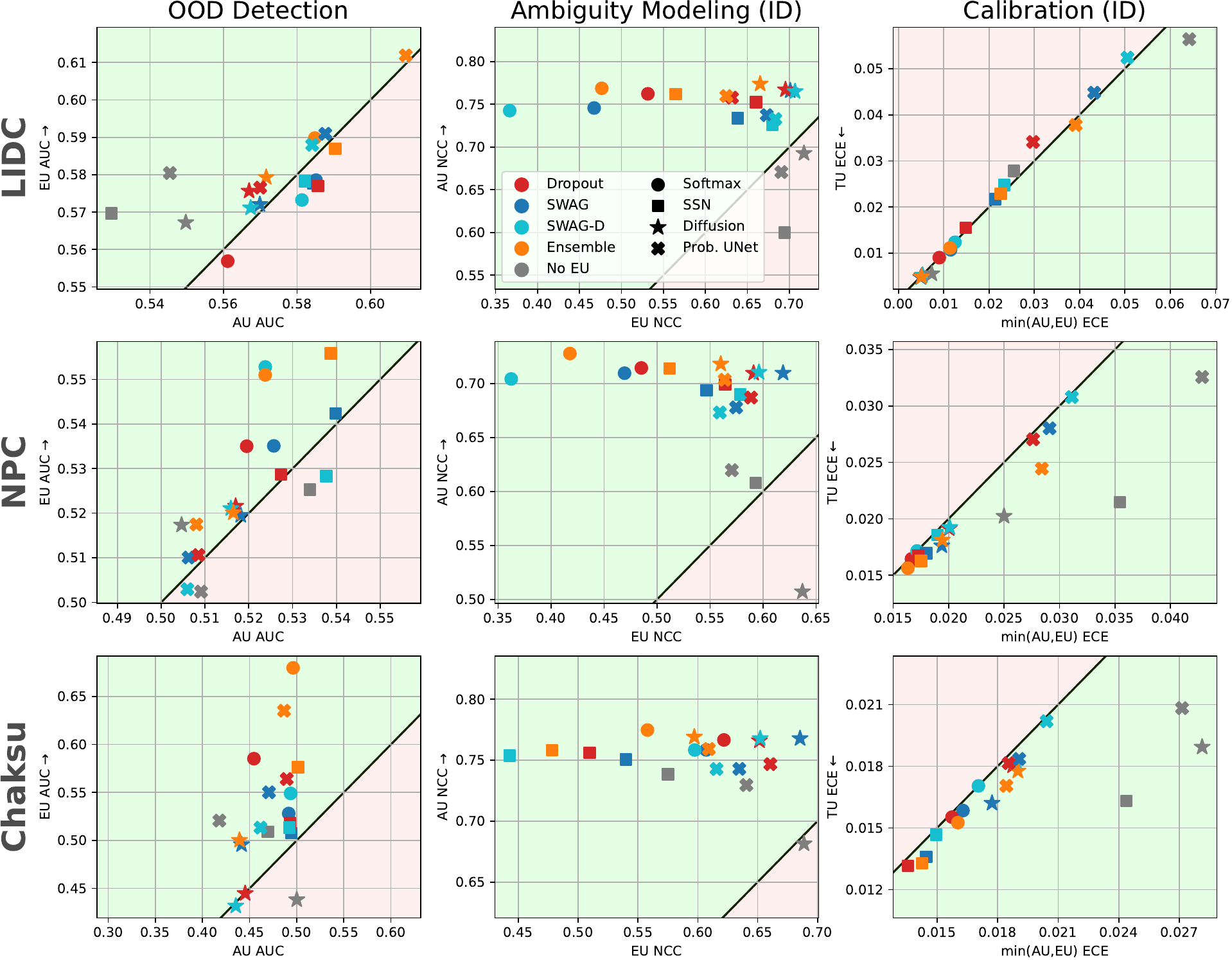}
    \caption{Scatter plots showing the performance of model combinations on the different tasks (OOD detection, AMB, CAL). Each point corresponds to a model combination, with y-values based on the theoretically consistent uncertainty measure and x-values using the inconsistent one.}
    \label{fig:scatter_grid_combined}
\end{figure}

\textbf{Experimental setup.}
We trained 19 valid AU-EU model combinations on LIDC, NPC and Chaksu. Predictions from the models (\cref{fig:pred_grid_id_npc}) are combined into uncertainty maps (\cref{fig:unc_grid_id_npc}) which we evaluate on the downstream tasks (OOD detection, AMB, CAL). OOD detection experiments use border-normalization as an aggregation strategy. Experiments were repeated 5 times with different seeds (mean values are shown in \cref{fig:scatter_grid_combined} and \cref{tab:ranking_tables}). For datasets with multiple augmented OOD test sets (LIDC and NPC), we evaluated each augmentation separately and report the mean across augmentation types. For convenience, a complete list of abbreviations is provided in \cref{supp:symbols_abbreviations}.

\begin{figure}[htbp]
    \centering
    \includegraphics[width=1.0\textwidth]{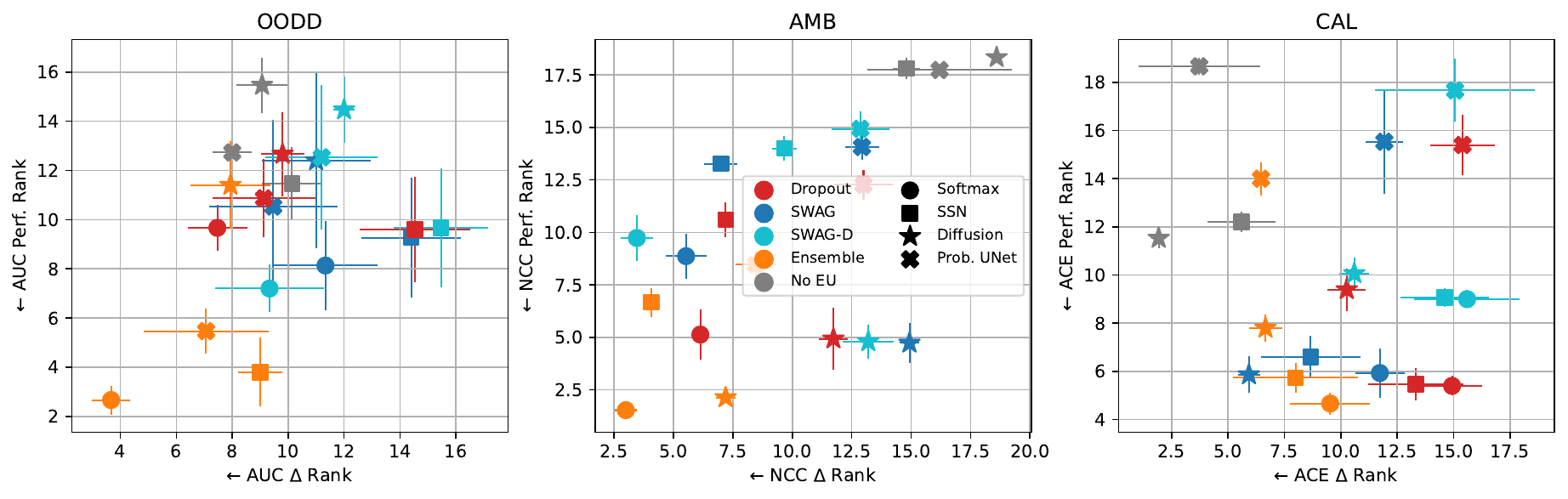}
    \caption{A scatter plot of the mean performance (aggregated over datasets) and entanglement-proxy ($\Delta$) rankings for different tasks. Lines indicate confidence intervals based on Student's $t$-distribution ($N=5$).}
    \label{fig:scatter_rank_task}
\end{figure}

\textbf{Performance and entanglement.}
Our central results on performance and the entanglement proxy can be seen in \cref{fig:scatter_grid_combined} as scatter plots in the coordinate system of $\Delta$. The y-axis shows the theoretically consistent uncertainty measure $(U_c)$, while the x-axis shows the theoretically inconsistent uncertainty measure $(U_i)$. Points below (OOD detection, AMB) or above (CAL) the $x=y$ line thus represent model combinations for which inconsistent uncertainty is better for a given task. We find that, as theoretically predicted, $U_c$ is better than $U_i$ for the majority of models and tasks. Ensembles perform remarkably well, both in task performance and in $\Delta$. Models with no EU component are located far from other models, usually with lower performance. Softmax and SSN models do well on both AMB and CAL performance while showing favorable $\Delta$ values.

\textbf{AU-EU combination effects.}
Under the assumption that there is no entanglement between the EU and AU estimations, OOD detection performance would be primarily determined by the EU model, AMB performance would be primarily determined by the AU model, and CAL performance would depend on both. This pattern is only matched to some degree in \cref{fig:scatter_grid_combined,fig:scatter_rank_task}. For AMB, AU model type affects $\Delta$ more than performance, and we observe that the points cluster based on AU (shape) rather than EU (color). SSN and softmax have more favorable entanglement-proxy values than Prob. UNet and Diffusion. For OOD detection, ensembles cluster far from other models. However, contrary to the theory and indicative of entanglement, the non-ensemble models still show an AU (shape) dependence by forming clusters (especially on LIDC data). Performance and $\Delta$ in CAL are largely determined by the AU (shape) model.

\textbf{Ranking overview.}
We rank all models by their consistent-metric performance (EU AUC, AU NCC, TU ECE) and by the entanglement proxy $\Delta$, averaging ranks across tasks. Results (\cref{tab:ranking_tables}) show that ensembles and softmax methods rank best overall. Additionally, diffusion models do well on LIDC and SSN models are decent on NPC and Chaksu.

\begin{table}[t]
\centering
\begin{subtable}{\textwidth}
\centering
\setlength{\tabcolsep}{2pt}
\resizebox{\textwidth}{!}{%
\begin{tabular}{ll rrrr p{8pt} rrrr p{8pt} rrrr}
    \multicolumn{2}{c}{\textbf{Perf. Rank}} & \multicolumn{14}{c}{\textbf{AU Method}} \\
    \cmidrule(lr){3-16}
    & & \multicolumn{1}{c}{\textbf{\shortstack{Soft-\\max}}}
      & \multicolumn{1}{c}{\textbf{SSN}}
      & \multicolumn{1}{c}{\textbf{\shortstack{Prob.\\UNet}}}
      & \multicolumn{1}{c}{\textbf{Diff.}}
      & 
      & \multicolumn{1}{c}{\textbf{\shortstack{Soft-\\max}}}
      & \multicolumn{1}{c}{\textbf{SSN}}
      & \multicolumn{1}{c}{\textbf{\shortstack{Prob.\\UNet}}}
      & \multicolumn{1}{c}{\textbf{Diff.}}
      & 
      & \multicolumn{1}{c}{\textbf{\shortstack{Soft-\\max}}}
      & \multicolumn{1}{c}{\textbf{SSN}}
      & \multicolumn{1}{c}{\textbf{\shortstack{Prob.\\UNet}}}
      & \multicolumn{1}{c}{\textbf{Diff.}} \\
    \midrule

    \multirow{5}{*}{\shortstack{\textbf{EU}\\\textbf{Method}}}
    & \textbf{No EU}    & \multicolumn{1}{c}{N/A} & 15.7 & 15.2 & 12.3 &   & \multicolumn{1}{c}{N/A} & 13.5 & 18.3 & 15.3 &   & \multicolumn{1}{c}{N/A} & 12.2 & 15.7 & 17.7 \\
    & \textbf{Dropout}  & 10.5 & 10.1 & 11.3 & 5.6 &   & 4.3 & 8.3 & 15.7 & 9.5 &   & 5.4 & 7.3 & 11.5 & 11.9 \\
    & \textbf{SWAG-D}   & 11.1 & 13.0 & 13.6 & 7.3 &   & 5.9 & 10.5 & 16.9 & 9.9 &   & 8.9 & 9.2 & 14.6 & 12.2 \\
    & \textbf{SWAG}     & 9.3 & 11.9 & 12.4 & 6.0 &   & 5.3 & 7.3 & 15.5 & 9.0 &   & 8.4 & 9.9 & 12.3 & 8.0 \\
    & \textbf{Ensemble} & 4.8 & 7.9 & 8.4 & 3.7 &   & 1.6 & 2.8 & 12.6 & 7.8 &   & 2.5 & 5.5 & 6.9 & 9.8 \\

    \midrule
    \\[-1.5em]
    \cmidrule(lr){3-6}\cmidrule(lr){8-11}\cmidrule(lr){13-16}
    \cmidrule(lr){3-6}\cmidrule(lr){8-11}\cmidrule(lr){13-16}

    % Dataset labels BELOW, with rules that stop exactly under their columns, skipping the ghost column.
    & & \multicolumn{4}{c}{\textbf{LIDC}} & & \multicolumn{4}{c}{\textbf{NPC}} & & \multicolumn{4}{c}{\textbf{Chaksu}} \\
\end{tabular}
}
\caption{Ranking of the model performance (EU AUC, AU NCC, TU ECE).}
\label{tab:performance_ranking}
\end{subtable}

\vspace{4pt}
\begin{subtable}{\textwidth}
\centering
\setlength{\tabcolsep}{2pt}
\resizebox{\textwidth}{!}{%
\begin{tabular}{ll rrrr p{8pt} rrrr p{8pt} rrrr}
    \multicolumn{2}{c}{\textbf{Ent. proxy ($\Delta$) Rank}} & \multicolumn{14}{c}{\textbf{AU Method}} \\
    \cmidrule(lr){3-16}
    & & \multicolumn{1}{c}{\textbf{\shortstack{Soft-\\max}}}
      & \multicolumn{1}{c}{\textbf{SSN}}
      & \multicolumn{1}{c}{\textbf{\shortstack{Prob.\\UNet}}}
      & \multicolumn{1}{c}{\textbf{Diff.}}
      & 
      & \multicolumn{1}{c}{\textbf{\shortstack{Soft-\\max}}}
      & \multicolumn{1}{c}{\textbf{SSN}}
      & \multicolumn{1}{c}{\textbf{\shortstack{Prob.\\UNet}}}
      & \multicolumn{1}{c}{\textbf{Diff.}}
      & 
      & \multicolumn{1}{c}{\textbf{\shortstack{Soft-\\max}}}
      & \multicolumn{1}{c}{\textbf{SSN}}
      & \multicolumn{1}{c}{\textbf{\shortstack{Prob.\\UNet}}}
      & \multicolumn{1}{c}{\textbf{Diff.}} \\
    \midrule

    \multirow{5}{*}{\shortstack{\textbf{EU}\\\textbf{Method}}}
    & \textbf{No EU}    & \multicolumn{1}{c}{N/A} & 11.5 & 8.3 & 7.6 &   & \multicolumn{1}{c}{N/A} & 12.3 & 12.1 & 8.8 &   & \multicolumn{1}{c}{N/A} & 6.7 & 7.5 & 13.2 \\
    & \textbf{Dropout}  & 9.9 & 13.1 & 10.5 & 8.1 &   & 8.4 & 12.0 & 14.1 & 10.1 &   & 10.2 & 9.9 & 13.0 & 13.6 \\
    & \textbf{SWAG-D}   & 9.1 & 14.2 & 11.7 & 11.5 &   & 7.1 & 15.3 & 14.5 & 11.1 &   & 12.3 & 10.3 & 12.9 & 13.3 \\
    & \textbf{SWAG}     & 8.3 & 12.1 & 11.4 & 10.2 &   & 7.9 & 9.0 & 11.9 & 11.2 &   & 12.3 & 8.9 & 11.0 & 10.5 \\
    & \textbf{Ensemble} & 6.4 & 10.3 & 9.1 & 6.5 &   & 4.8 & 5.1 & 6.7 & 7.7 &   & 5.0 & 5.6 & 6.2 & 7.5 \\

    \midrule
    \\[-1.5em]
    \cmidrule(lr){3-6}\cmidrule(lr){8-11}\cmidrule(lr){13-16}
    \cmidrule(lr){3-6}\cmidrule(lr){8-11}\cmidrule(lr){13-16}

    % Dataset labels BELOW, with rules that stop exactly under their columns, skipping the ghost column.
    & & \multicolumn{4}{c}{\textbf{LIDC}} & & \multicolumn{4}{c}{\textbf{NPC}} & & \multicolumn{4}{c}{\textbf{Chaksu}} \\
\end{tabular}
}
\caption{Ranking by the entanglement proxy ($\Delta$).}
\label{tab:entanglement_ranking}
\end{subtable}

\caption{Average ranking (1-19) of performance and the entanglement proxy ($\Delta$) when comparing the 19 AU-EU model combinations. The ranks were calculated separately per task and averaged across tasks.}
\label{tab:ranking_tables}
\end{table}

\begin{figure}[ht]
    \centering
    \includegraphics[width=\textwidth]{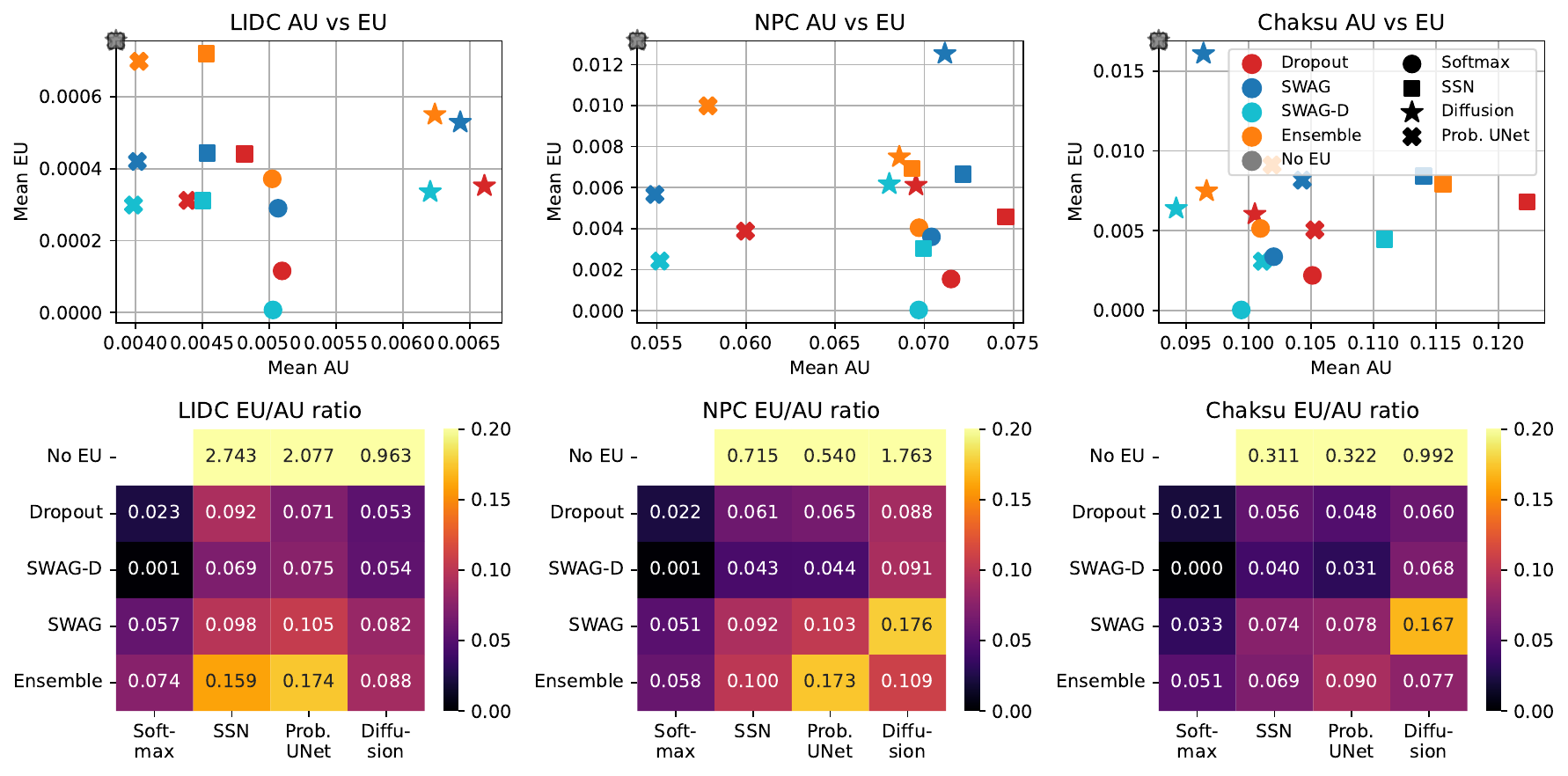}
    \caption{\textbf{Top}: Scatter plots comparing mean AU and EU across models and datasets. \textbf{Bottom}: Mean ratio of EU/AU across models and datasets on OOD images (using image-wise mean). No EU models are left out of axis limits for better resolution.}
    \label{fig:eu_collapse}
\end{figure}

\textbf{Epistemic collapse.}
Epistemic collapse causes EU component models to mostly predict the same generative mean. As a result, the magnitude of EU maps are far smaller than AU maps, meaning $TU\approx AU$, and causing entanglement between TU and AU. This can be observed in \cref{fig:scatter_grid_combined} with models being close to the diagonal. With the exception of the "no EU" models, the mean EU/AU ratio is always less than 0.2 (see \cref{fig:eu_collapse}). The problem is more severe for dropout and SWAG-D than SWAG and ensembles. Softmax models are significantly more affected by epistemic collapse than generative models.

\begin{figure}[ht]
    \centering
    \includegraphics[width=1.0\textwidth]{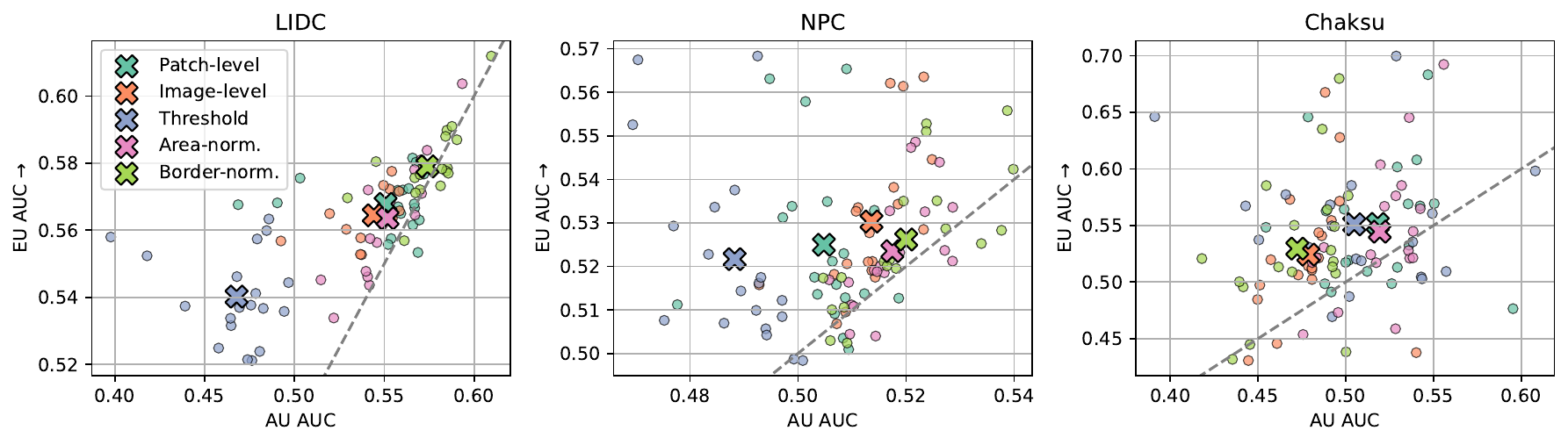}
    \caption{Scatter plots showing OOD detection performance as different aggregation strategies are used. Each small point is an AU-EU model combination, and the large crosses are the mean performances across all combinations.}
    \label{fig:scatter_aggr}
\end{figure}

\textbf{OOD detection aggregation strategy.}
We ablate the aggregation strategy for OOD detection (\cref{fig:scatter_aggr}), finding it matters mostly for LIDC, but less for Chaksu and NPC. Chaksu models cluster similarly across strategies, likely because the dataset is more class-balanced with no empty labels. For LIDC, border-normalization outperforms alternatives while threshold-based normalization performs worst. Methods accounting for mask size variation do better.

\begin{table}[htbp]
% Define cell dimensions
\def\cellw{3.0}      % cell width (all columns)
\def\cellh{0.8}      % standard row height (all rows)
\def\cellgap{0.08}   % gap between cells
\def\cellgaphalf{0.04} % half of the gap, for convenience
% Command for all cells with optional second color for lower-right triangle
\newcommand{\tikzmakecell}[4][]{
  \fill[#3] (#2) rectangle ++(\cellw,\cellh);
  \if!#1!\else\fill[#1] (#2) -- ++({\cellw},0) -- ++(0,\cellh) -- cycle;\fi
  \node[text width=\cellw cm-0.2cm,align=center] at ([shift={(\cellw/2,\cellh/2)}]#2) {\vphantom{A}#4\vphantom{g}};
}
    
\centering
\resizebox{\textwidth}{!}{%
\begin{tikzpicture}
  % Define grid positions 
  \foreach \col [count=\x from 0] in {A,B,C,D} {
    \pgfmathsetmacro{\xpos}{\x*\cellw + \x*\cellgap}
    \foreach \row [count=\y from 0] in {1,2,3,4,5} {
      \pgfmathsetmacro{\ypos}{-\y*\cellh - \y*\cellgap}
      \coordinate (\col\row) at (\xpos,\ypos);
    }
  }
  
  % Place cells using \tikzmakecell{position}{color}{text}

  % Header row
  \def\curcellh{\cellh}
  \tikzmakecell{A1}{white}{}
  \tikzmakecell{B1}{white}{\tblsmall{\\[0.5ex]Performance}}
  \tikzmakecell{C1}{white}{\tblsmall{\\[0.5ex]Ent. proxy ($\Delta$)}}
  \tikzmakecell{D1}{white}{\tblsmall{\\[0.5ex]Both}}
  
  % Row 2
  \def\curcellh{\cellh}
  \tikzmakecell{A2}{white}{\tblsmall{Out-of-distribution\\detection}}
  \tikzmakecell[myred]{B2}{mygreen}{SSN ens.\\/Softmax ens.}
  \tikzmakecell{C2}{myred}{Softmax ens.}
  \tikzmakecell{D2}{myred}{Softmax ens.}
  
  % Row 3
  \def\curcellh{\smallcellh}
  \tikzmakecell{A3}{white}{\tblsmall{Amb. Mod. (AMB)}}
  \tikzmakecell[myred]{B3}{myblue}{Diffusion ens.\\/Softmax ens.}
  \tikzmakecell[myred]{C3}{mygreen}{SSN ens.\\/Softmax ens.}
  \tikzmakecell{D3}{myred}{Softmax ens.}
  % Row 4: reduced height
  \def\curcellh{\smallcellh}
  \tikzmakecell{A4}{white}{\tblsmall{Calibration (CAL)}}
  \tikzmakecell{B4}{myred}{Softmax ens.}
  \tikzmakecell{C4}{mymagenta}{Diffusion SWAG}
  \tikzmakecell[myred]{D4}{mymagenta}{Diffusion SWAG\\/Softmax ens.}
  
  % Row 5: reduced height
  \def\curcellh{\smallcellh}
  \tikzmakecell{A5}{white}{\tblsmall{Overall}}
  \tikzmakecell{B5}{myred}{Softmax ens.}
  \tikzmakecell{C5}{myred}{Softmax ens.}
  \tikzmakecell{D5}{myred}{Softmax ens.}
  % Total table width
\pgfmathsetmacro{\tablewidth}{4*\cellw + 3*\cellgap}

% Top rule
\draw[line width=0.6pt] (-\cellgap,\cellh-0.35) -- (\tablewidth+\cellgap,\cellh-0.35);

% Rule after header row
\draw[line width=0.4pt] (0,-\cellgaphalf) -- (\tablewidth,-\cellgaphalf);

% Bottom rule
\draw[line width=0.6pt] (-\cellgap,-5*\cellh+0.43) -- (\tablewidth+\cellgap,-5*\cellh+0.43);
\end{tikzpicture}
}
\caption{Our recommendations for AU-EU model combinations across tasks, based on whether performance, favorable entanglement-proxy values or both are important. AU-EU model combinations are assigned colors for an easy overview.}
\label{tab:recommendation_table}
\end{table}

%% file: text/5discussion.tex
\section{Discussion}

\textbf{Epistemic collapse as a source of entanglement.}
EU magnitudes are substantially smaller than AU for nearly all models (epistemic collapse). This is especially problematic for CAL, where under the Kendall \& Gal decomposition $TU = AU+EU \approx AU$ prevents TU from providing information beyond AU. Addressing epistemic collapse could therefore affect $\Delta$ in CAL. Kirsch et al.~\cite{implicit_ensemble_collapse} proposed extracting independent sub-networks post-training, and incorporating their methods in our framework is a promising way forward. We tried a post-hoc correction to boost EU (\eg by multiplication with a constant) before computing TU, thereby making the two components more comparable in magnitude, however this did not work empirically and we think epistemic collapse needs to be addressed on a fundamental level to reduce entanglement.

\textbf{Low entanglement in ensembles.}
Ensembles exhibit the best performance and most favorable entanglement-proxy values across most tasks. A plausible explanation is that each ensemble member follows an independent training trajectory from a different random initialization, leading to genuinely diverse network weights and predictions. Dropout and SWAG/SWAG-D obtain weight distributions from a single training run, constraining diversity and potentially causing EU to capture less epistemic variation, which leaks into AU. We observe less severe epistemic collapse for ensembles, supporting this.

\textbf{Models without an EU component.}
No EU models are consistently outliers in \cref{fig:scatter_grid_combined}: their proxy values are less favorable in CAL and more favorable in AMB, with generally worse performance. Since these models have no EU component and their predictions are used directly as the generative mean, epistemic collapse does not apply, explaining their lower CAL proxy values. Otherwise, these models show no clear trends consistent with a meaningful AU-EU decomposition, even if we swap AU and EU labels as ValUES~\cite{ValUES}. We therefore argue against applying the Kendall \& Gal decomposition to models without an explicit EU component, as done in ValUES~\cite{ValUES}. To our knowledge, this application lacks a formal theoretical foundation in existing literature, and we therefore caution against it.

\textbf{Model choice recommendations.}
Model selection involves trading off performance, the entanglement proxy and computational cost. Ensembles require \eg $N=5$ model trainings instead of 1, while dropout and SWAG/SWAG-D only train once, with SWAG needing more storage. Among non-ensemble EU methods, dropout performs slightly better overall. If ensembles are too expensive given their training compute scaling with ensemble size, dropout could be the preferred cheaper option. For OOD detection, ensembles excel, with the best performance and most favorable proxy values. For AMB, softmax and SSN have more favorable $\Delta$ values. For CAL, softmax and SSN perform best while Prob.\ UNet has the most favorable proxy value. Overall, a softmax ensemble performs surprisingly well given its conceptual simplicity. Ignoring compute budget and ease of implementation, we outline our model recommendations in \cref{tab:recommendation_table}.

\textbf{Limitations and future directions.}
Our study has several limitations. Tracking how entanglement evolves during training could shed light on when and why it emerges. Studying entanglement as a function of model size would clarify the role of epistemic collapse, which worsens with increasing capacity~\cite{implicit_ensemble_collapse,uq_hole_collapse}. Incorporating non-generative AU methods such as test-time augmentation~\cite{ayhan2018test} would test whether trends generalize beyond probabilistic models.
We also omit feature- and logit-based OOD baselines such as energy scores, Mahalanobis distance, and ViM as well as evidential deep learning, which do not map directly to the probability-based AU/EU/TU framework used here. Comparing a broader class of models is an important direction for future work.
Further, our evaluation is limited to three medical imaging datasets, each with its own flaws. The aleatoric uncertainty in LIDC is not ideal since nodules are either present or absent, with little room for ambiguous mask delineations to have an effect. The LIDC and NPC datasets are natively 3D, but we sliced them into 2D images. While volumetric evaluation would be preferable, other native 2D datasets were too small to replace these data, and 2D LIDC/NPC slices are widely used in prior work. All datasets are cropped around the region of interest, which makes segmentation unnaturally easy in some applications. These choices were motivated by common practice for LIDC, improved segmentation fidelity for Chaksu (the optic disc is relatively easy to identify but small compared to the retina), and square cropping for NPC (without centralization). We think going beyond $128\times128$ resolution would not materially change downstream trends, since early experiments at $64\times64$ resolution were similar, and all our datasets were sampled close to native quality at $128\times128$ resolution. The ValUES framework requires datasets with multiple expert annotations, which are scarce due to the large annotation burden. As a result, most such datasets are relatively small or cover narrow segmentation problems with few structures. Finally, our study lacks large-scale clinical validation. Future work should acquire high-quality multi-annotator datasets and validate findings in broader clinical settings.

\section{Conclusion}
Our systematic evaluation of 19 AU-EU model combinations on 3 datasets has shed light on the usefulness of these approaches and the challenges in uncertainty quantification. Our proposed entanglement proxy summarizes whether the theoretically consistent uncertainty measure outperforms the inconsistent one. Deep ensembles consistently achieve the best performance and most favorable proxy values. We identified epistemic collapse as a possible driver of entanglement and suggested potential mitigation strategies. Our results show that applying the Kendall \& Gal decomposition to models without an explicit EU component is not well-motivated. Unless a generative model is required, our findings suggest that an ensemble of standard cross-entropy-trained softmax models is sufficient for downstream task performance. Our benchmark and entanglement proxy provide the tools to diagnose and reduce entanglement in future methods and ultimately develop better uncertainty quantification methods.

\section{Acknowledgements}
This work was supported by Danish Data Science Academy, which is funded by the Novo Nordisk Foundation (NNF21SA0069429) and VILLUM FONDEN (40516).
\newpage

%% file: text/6supplement.tex
\clearpage  % Start supplementary material on a new page
\appendix
\section{Supplementary Materials}

To supplement the paper's findings, we have added qualitative plots showing model and dataset results, along with details on the model architecture and predictions. Due to space limits, we could not accommodate these in the main paper.

\subsection{Abbreviations Overview}
\label{supp:symbols_abbreviations}

\begin{table}[h]
\centering
\caption{Abbreviations used throughout the paper.}
\label{tab:abbreviations}
\begin{tabular}{ll}
\toprule
Abbreviation & Meaning \\
\midrule
AU & Aleatoric uncertainty \\
EU & Epistemic uncertainty \\
TU & Total uncertainty \\
ID & In-distribution \\
OOD & Out-of-distribution \\
AMB & Ambiguity modeling \\
CAL & Calibration \\
ECE & Expected calibration error \\
AUROC/AUC & Area under the receiver operating characteristic curve \\
NCC & Normalized cross-correlation \\
SSN & Stochastic Segmentation Network \\
Prob. UNet & Probabilistic UNet \\
MC Dropout & Monte Carlo dropout \\
SWAG & Stochastic weight averaging-Gaussian \\
SWAG-D & Diagonal-only SWAG \\
LIDC & LIDC-IDRI dataset \\
NPC & MMIS NPC-170 dataset \\
Chaksu & Ch\'{a}k\c{s}u IMAGE dataset \\
CT & Computed tomography \\
MRI & Magnetic resonance imaging \\
\bottomrule
\end{tabular}
\end{table}

\clearpage
\subsection{Additional Qualitative Plots}
\FloatBarrier
\begin{figure}[htb]
    \centering
    \begin{subfigure}[htbp]{\textwidth}
        \centering
        \includegraphics[width=\textwidth]{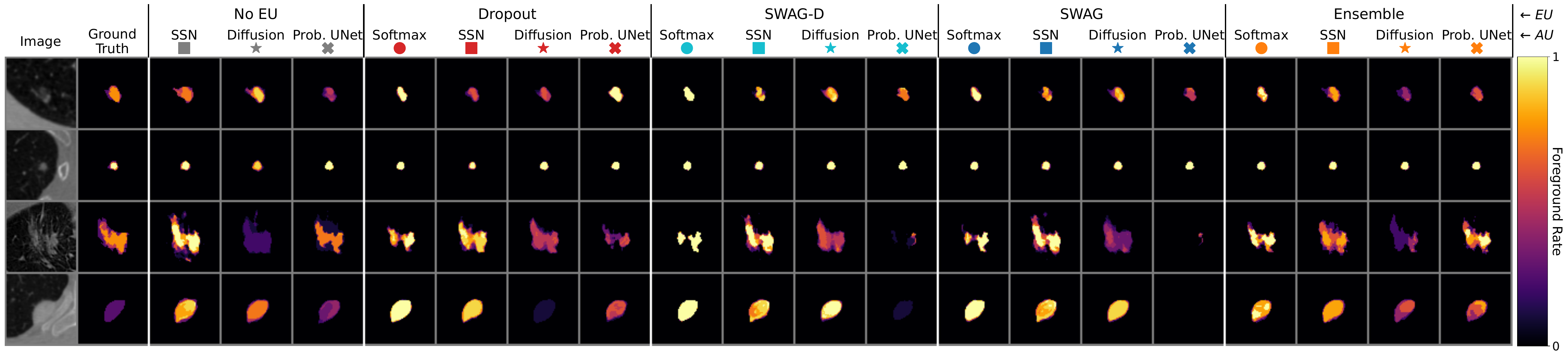}
        \caption{LIDC dataset (ID).}
        \label{fig:qual_grid_id_lidc}
    \end{subfigure}
    \begin{subfigure}[htbp]{\textwidth}
        \centering
        \includegraphics[width=\textwidth]{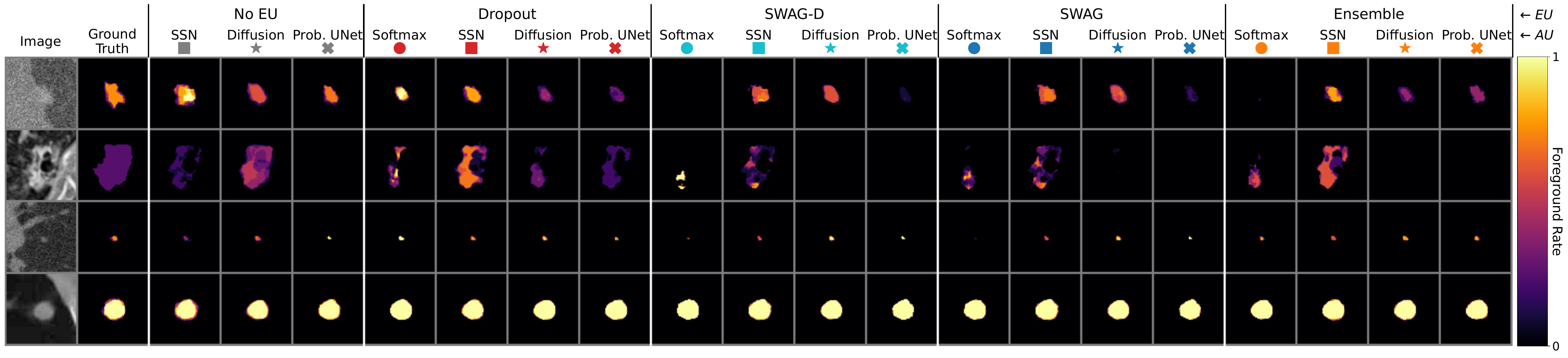}
        \caption{LIDC dataset (OOD).}
        \label{fig:qual_grid_ood_lidc}
    \end{subfigure}
    \begin{subfigure}[htbp]{\textwidth}
        \centering
        \includegraphics[width=\textwidth]{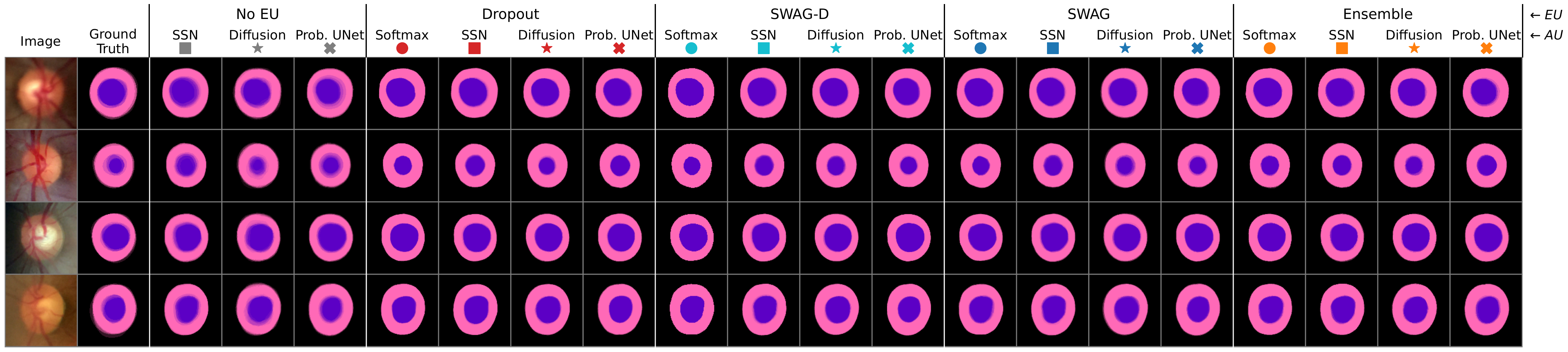}
        \caption{Chaksu dataset (ID). Purple, pink and black represents cup, disc and background, respectively.}
        \label{fig:qual_grid_id_chaksu}
    \end{subfigure}
    \begin{subfigure}[htbp]{\textwidth}
        \centering
        \includegraphics[width=\textwidth]{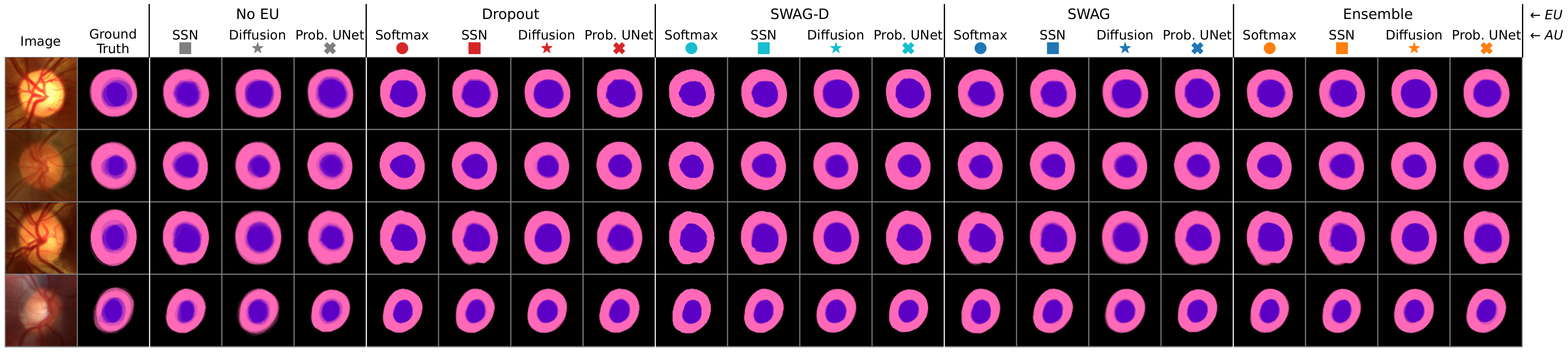}
        \caption{Chaksu dataset (OOD). Purple, pink and black represent the cup, disc and background, respectively.}
        \label{fig:qual_grid_ood_chaksu}
    \end{subfigure}

    \caption{Mean predictions ($\mathbb{E}_\theta[\mathbb{E}_y[p]]$) for the different model combinations across datasets and distribution settings.}
    \label{fig:qual_grid_combined}
\end{figure}

\begin{figure}[htbp]
    \centering
    \includegraphics[width=\textwidth]{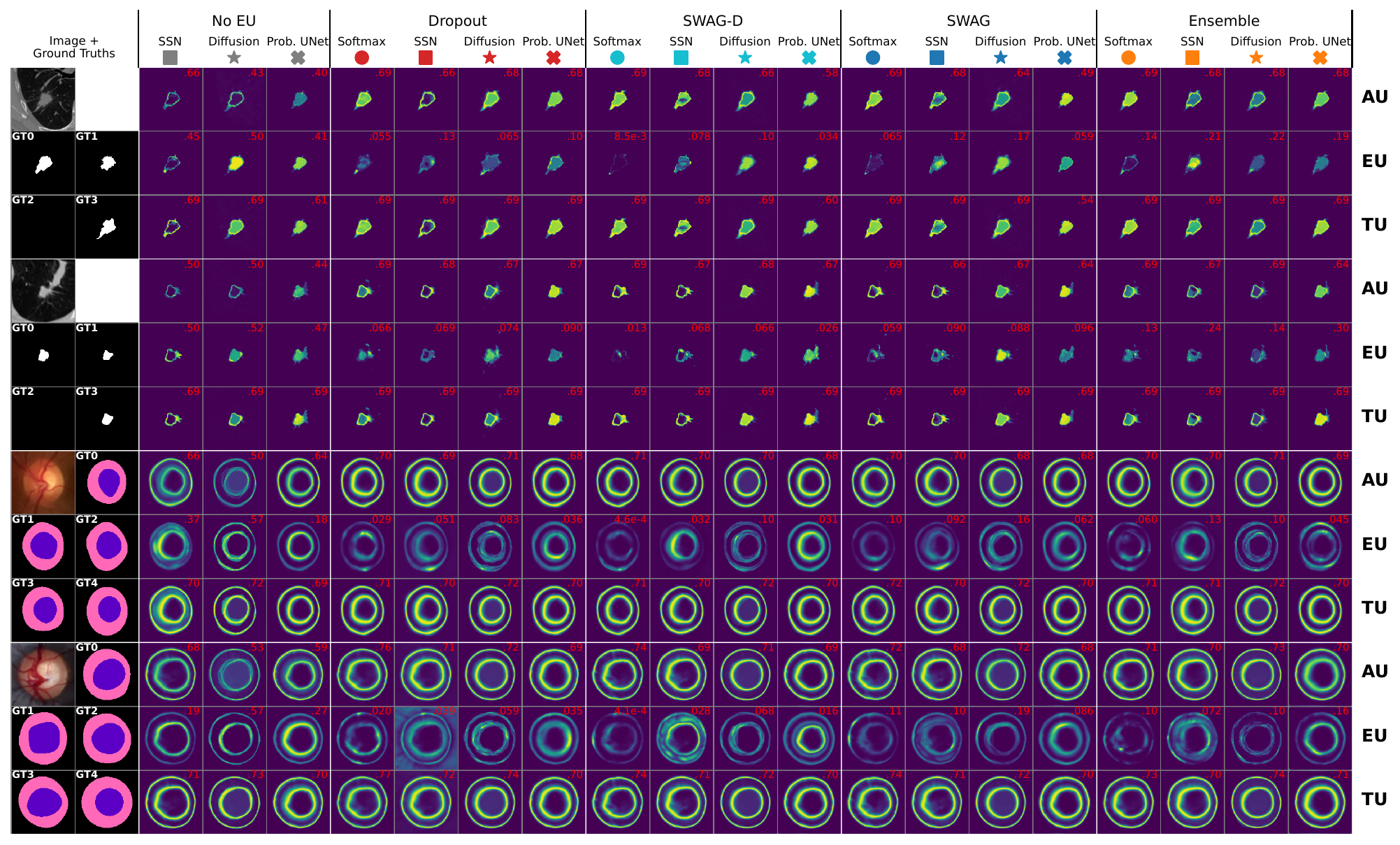}
    \caption{Aleatoric (AU), epistemic (EU) and total uncertainty (TU) maps for in-distribution (ID) images from the LIDC and Chaksu datasets. Red numbers indicate the maximum value.
    }
    \label{fig:qual_grid_unc_id}
\end{figure}

\begin{figure}[htbp]
    \centering
    \includegraphics[width=\textwidth]{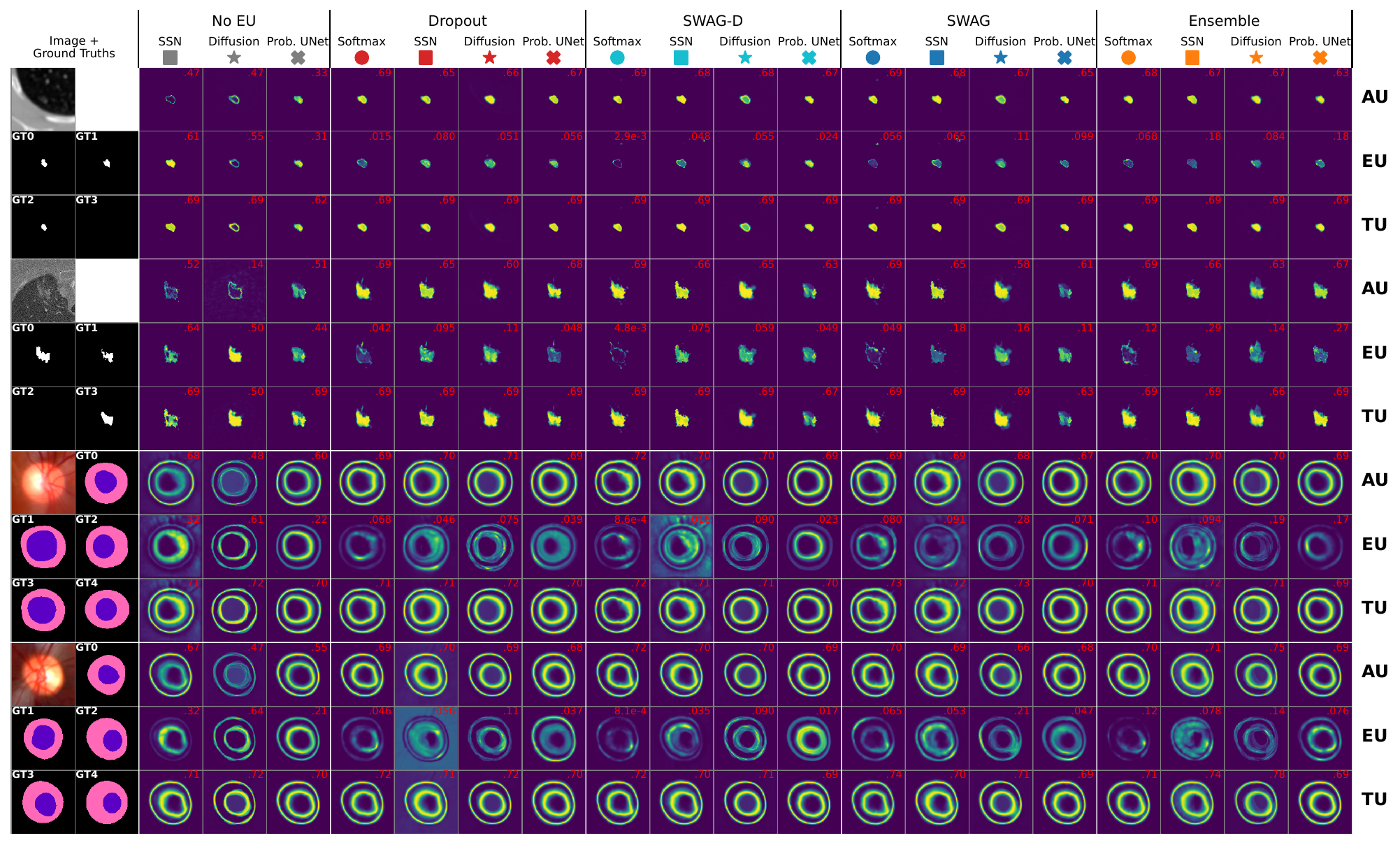}
    \caption{Aleatoric (AU), epistemic (EU) and total uncertainty (TU) maps for out-of-distribution (OOD) images from the LIDC and Chaksu datasets. Red numbers indicate the maximum value.    }
    \label{fig:qual_grid_unc_ood}
\end{figure}

\FloatBarrier
\subsection{Architecture Details}

The UNet architecture uses 32 base channels which are increased with a channel multiplier of $(1,2,4,8)$ across the different resolution scales. The backbone mainly consists of residual blocks (ResBlocks) which use the following layers:
\begin{itemize}
    \item \textbf{Norm}: Group normalization with 32 channels and $\epsilon=$1e-5.
    \item \textbf{SiLU}: The Sigmoid Linear Unit, $\text{SiLU}(x)=x/(1+e^{-x})$ (also used as the activation function throughout the network)
    \item \textbf{Conv}: A standard 2D convolution with kernel size $3\times3$.
\end{itemize}
A ResBlock is defined by sequential application of the layers: Norm, SiLU, Conv, Norm, SiLU, Conv. The ResBlocks are residual, meaning the input is added to the output. If the number of channels is changed through a ResBlock, a $1\times1$ 2D convolution is added to the residual path to account for the change. We use 3 downscaling/upscaling operations (as implied by the channel multiplier). Downscaling uses average pooling and upscaling uses nearest-neighbour interpolation followed by a $3\times3$ 2D convolution. Each resolution scale has 2 ResBlocks in succession, with an additional 2 middle ResBlocks. We use additive skip connections between the encoder and decoder just before/after layers with downscaling/upscaling. The number of channels is changed in the first convolution after a downscaling/upscaling operation. Self-attention is used after each ResBlock when the feature map has a width of 32 or less (the two deepest scales). When applying self-attention, we consider each pixel\footnote{Pixel is used loosely here, referring to a spatial position in the downscaled feature tensor.} as a separate token. 

To enable reproducibility and specify implementation details not covered in the text, the code can be viewed at \texttt{github.com/anonymous}.

\FloatBarrier
\subsection{Discretizing Predictions}
One may notice that model predictions in \cref{fig:pred_grid_intro} are blurry around the edges. This reflects an intentional design choice, as we decided not to discretize model predictions. Model predictions were represented as probability maps produced by a softmax layer. Early experiments showed that converting these probability maps to discrete one-hot maps using the most probable class would either hurt uncertainty quantification performance or make no difference.
\newpage
\FloatBarrier

\subsection{Additional Scatter Plots}
\begin{figure}[htbp]
    \centering
    \includegraphics[width=1.0\textwidth]{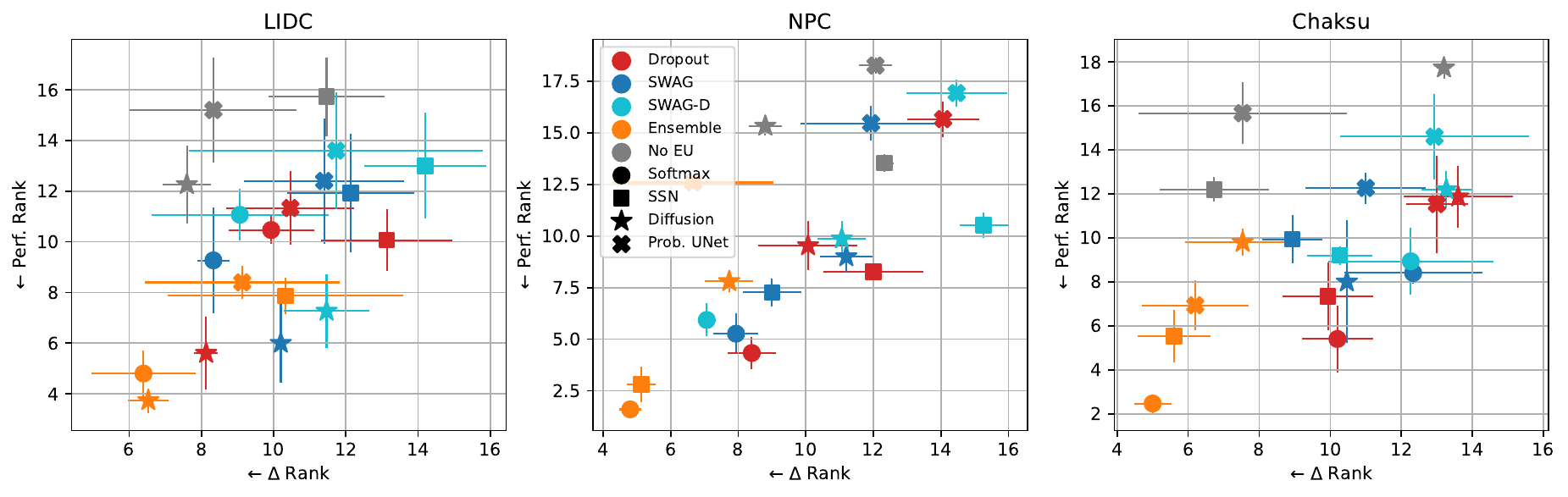}
    \caption{Mean performance and entanglement-proxy ($\Delta$) ranking for different datasets (aggregated over tasks). Lines indicate confidence intervals based on Student's $t$-distribution ($N=5$).}
    \label{fig:scatter_rank}
\end{figure}

\begin{figure}[!htbp]
    \centering
    \includegraphics[width=\textwidth]{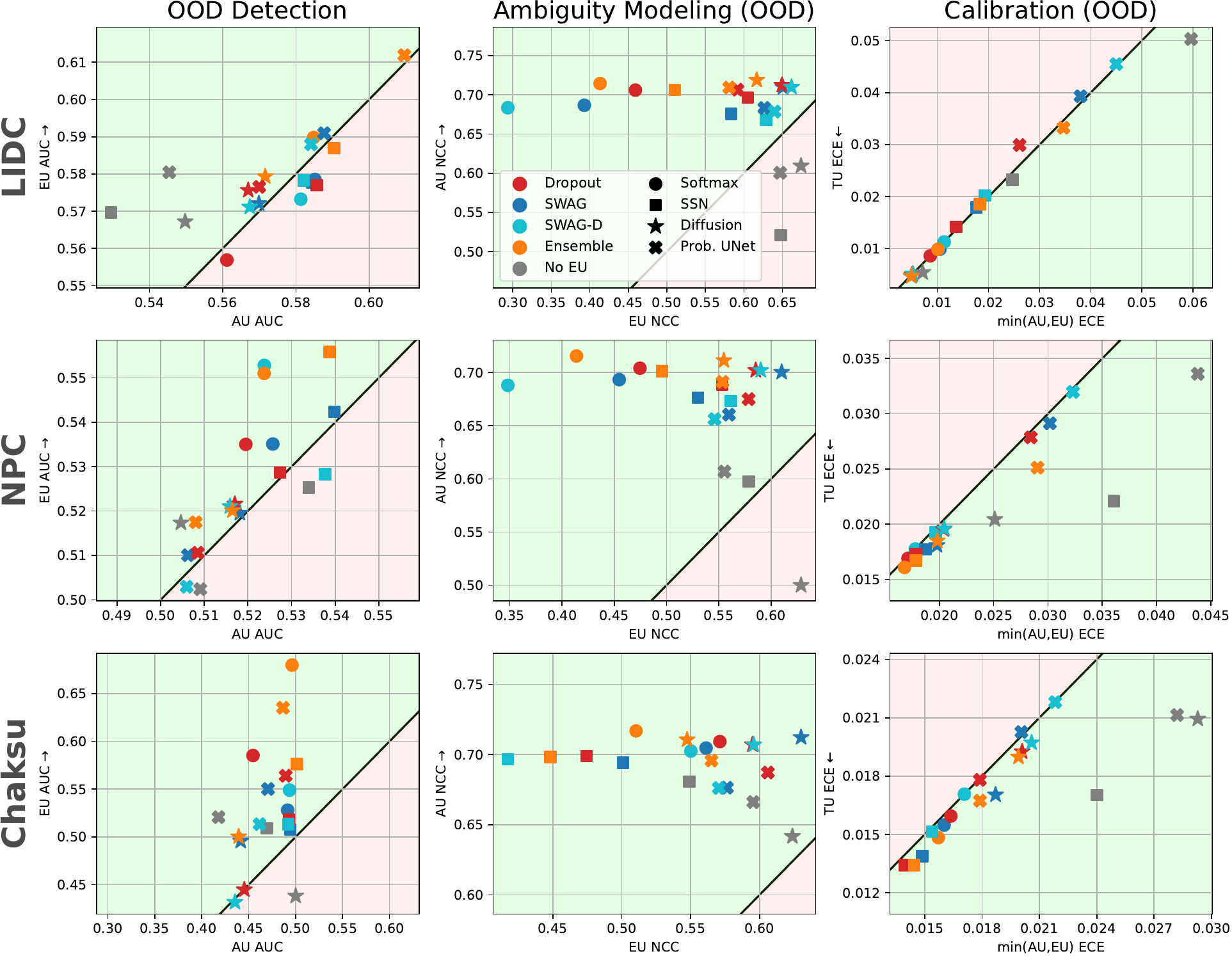}
    \caption{Identical to \cref{fig:scatter_grid_combined} but with OOD splits for AMB and CAL.}
    \label{fig:scatter_grid_combined_OOD}
\end{figure}

\begin{figure}[!htbp]
    \centering
    \includegraphics[width=\textwidth]{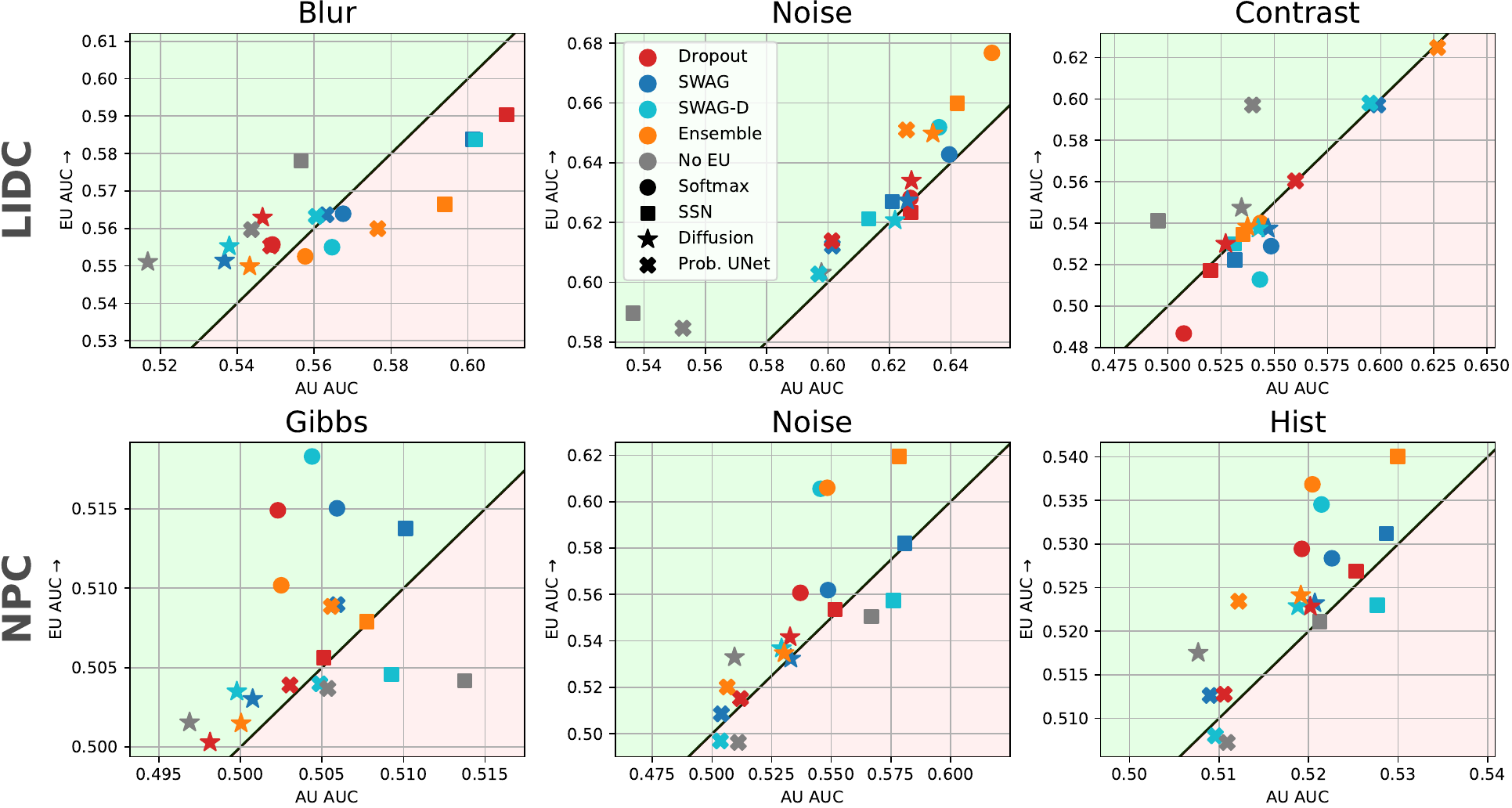}
    \caption{Same as the left column of \cref{fig:scatter_grid_combined} but with each OOD split for LIDC (blur, noise, contrast) and NPC (Gibbs, noise, hist) shown separately.}
    \label{fig:scatter_grid_combined_OOD_split}
\end{figure}

\clearpage
\subsection{Influence of the AU-component Sample Count}
\begin{figure}[H]
    \centering
    \includegraphics[width=0.93\textwidth]{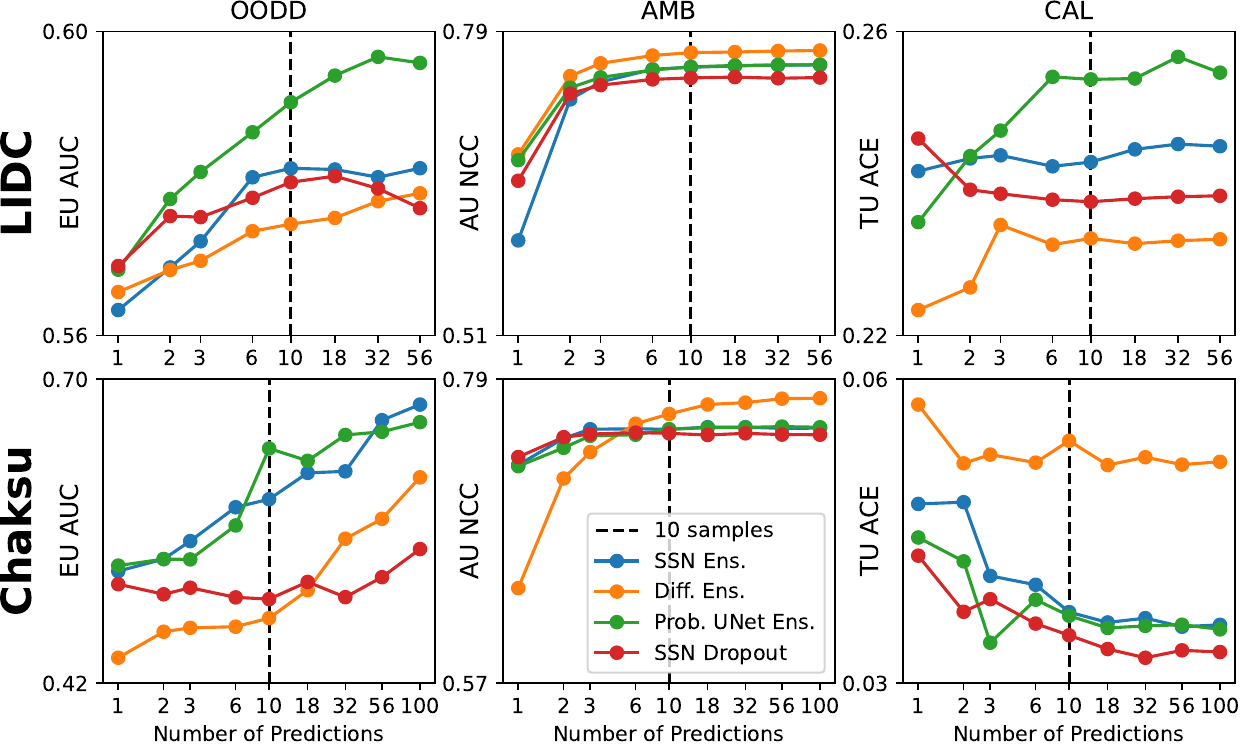}
    \caption{The performance metrics as the number of AU predictions are varied. We used 10 samples in the paper as a performance vs compute trade-off.}
    \label{fig:n_pred_plot}
\end{figure}

\subsection{Results with Confidence Intervals}

\begin{table}[H]
\centering
\begin{subtable}{\textwidth}
\centering
\setlength{\tabcolsep}{2pt}
\resizebox{\textwidth}{!}{%
\begin{tabular}{ll rrrr p{8pt} rrrr p{8pt} rrrr}
    \multicolumn{2}{c}{\textbf{Perf. Rank}} & \multicolumn{14}{c}{\textbf{AU Method}} \\
    \cmidrule(lr){3-16}
    & & \multicolumn{1}{c}{\textbf{\shortstack{Soft-\\max}}}
      & \multicolumn{1}{c}{\textbf{SSN}}
      & \multicolumn{1}{c}{\textbf{\shortstack{Prob.\\UNet}}}
      & \multicolumn{1}{c}{\textbf{Diff.}}
      & 
      & \multicolumn{1}{c}{\textbf{\shortstack{Soft-\\max}}}
      & \multicolumn{1}{c}{\textbf{SSN}}
      & \multicolumn{1}{c}{\textbf{\shortstack{Prob.\\UNet}}}
      & \multicolumn{1}{c}{\textbf{Diff.}}
      & 
      & \multicolumn{1}{c}{\textbf{\shortstack{Soft-\\max}}}
      & \multicolumn{1}{c}{\textbf{SSN}}
      & \multicolumn{1}{c}{\textbf{\shortstack{Prob.\\UNet}}}
      & \multicolumn{1}{c}{\textbf{Diff.}} \\
    \midrule

    \multirow{5}{*}{\shortstack{\textbf{EU}\\\textbf{Method}}}
    & \textbf{No EU}    & \multicolumn{1}{c}{N/A} & 15.7±1.7 & 15.2±2.3 & 12.3±1.7 &   & \multicolumn{1}{c}{N/A} & 13.5±0.5 & 18.3±0.5 & 15.3±0.5 &   & \multicolumn{1}{c}{N/A} & 12.2±0.6 & 15.7±1.6 & 17.7±0.5 \\
    & \textbf{Dropout}  & 10.5±0.6 & 10.1±1.4 & 11.3±1.6 & 5.6±1.6 &   & 4.3±0.9 & 8.3±1.4 & 15.7±1.0 & 9.5±1.3 &   & 5.4±1.7 & 7.3±1.7 & 11.5±2.5 & 11.9±1.6 \\
    & \textbf{SWAG-D}   & 11.1±1.1 & 13.0±2.3 & 13.6±2.6 & 7.3±1.6 &   & 5.9±0.9 & 10.5±0.7 & 16.9±0.7 & 9.9±1.0 &   & 8.9±1.7 & 9.2±0.5 & 14.6±2.2 & 12.2±1.0 \\
    & \textbf{SWAG}     & 9.3±2.3 & 11.9±2.6 & 12.4±2.8 & 6.0±1.7 &   & 5.3±1.1 & 7.3±0.7 & 15.5±1.0 & 9.0±0.9 &   & 8.4±0.5 & 9.9±1.2 & 12.3±0.8 & 8.0±3.1 \\
    & \textbf{Ensemble} & 4.8±1.0 & 7.9±0.8 & 8.4±0.7 & 3.7±0.5 &   & 1.6±0.2 & 2.8±1.0 & 12.6±0.3 & 7.8±0.6 &   & 2.5±0.5 & 5.5±1.3 & 6.9±1.3 & 9.8±0.7 \\

    \midrule
    \\[-1.5em]
    \cmidrule(lr){3-6}\cmidrule(lr){8-11}\cmidrule(lr){13-16}
    \cmidrule(lr){3-6}\cmidrule(lr){8-11}\cmidrule(lr){13-16}

    % Dataset labels BELOW, with rules that stop exactly under their columns, skipping the ghost column.
    & & \multicolumn{4}{c}{\textbf{LIDC}} & & \multicolumn{4}{c}{\textbf{NPC}} & & \multicolumn{4}{c}{\textbf{Chaksu}} \\
\end{tabular}
}
\caption{Ranking of the model performance (EU AUC, AU NCC, TU ECE).}
\label{tab:performance_ranking_conf}
\end{subtable}

\vspace{4pt}
\begin{subtable}{\textwidth}
\centering
\setlength{\tabcolsep}{2pt}
\resizebox{\textwidth}{!}{%
\begin{tabular}{ll rrrr p{8pt} rrrr p{8pt} rrrr}
    \multicolumn{2}{c}{\textbf{Ent. proxy ($\Delta$) Rank}} & \multicolumn{14}{c}{\textbf{AU Method}} \\
    \cmidrule(lr){3-16}
    & & \multicolumn{1}{c}{\textbf{\shortstack{Soft-\\max}}}
      & \multicolumn{1}{c}{\textbf{SSN}}
      & \multicolumn{1}{c}{\textbf{\shortstack{Prob.\\UNet}}}
      & \multicolumn{1}{c}{\textbf{Diff.}}
      & 
      & \multicolumn{1}{c}{\textbf{\shortstack{Soft-\\max}}}
      & \multicolumn{1}{c}{\textbf{SSN}}
      & \multicolumn{1}{c}{\textbf{\shortstack{Prob.\\UNet}}}
      & \multicolumn{1}{c}{\textbf{Diff.}}
      & 
      & \multicolumn{1}{c}{\textbf{\shortstack{Soft-\\max}}}
      & \multicolumn{1}{c}{\textbf{SSN}}
      & \multicolumn{1}{c}{\textbf{\shortstack{Prob.\\UNet}}}
      & \multicolumn{1}{c}{\textbf{Diff.}} \\
    \midrule

    \multirow{5}{*}{\shortstack{\textbf{EU}\\\textbf{Method}}}
    & \textbf{No EU}    & \multicolumn{1}{c}{N/A} & 11.5±1.8 & 8.3±2.6 & 7.6±0.7 &   & \multicolumn{1}{c}{N/A} & 12.3±0.3 & 12.1±0.5 & 8.8±0.6 &   & \multicolumn{1}{c}{N/A} & 6.7±1.7 & 7.5±3.3 & 13.2±0.2 \\
    & \textbf{Dropout}  & 9.9±1.3 & 13.1±2.0 & 10.5±2.0 & 8.1±0.4 &   & 8.4±0.8 & 12.0±1.7 & 14.1±1.2 & 10.1±1.6 &   & 10.2±1.1 & 9.9±1.4 & 13.0±1.0 & 13.6±1.7 \\
    & \textbf{SWAG-D}   & 9.1±2.7 & 14.2±1.9 & 11.7±4.5 & 11.5±1.3 &   & 7.1±0.2 & 15.3±0.8 & 14.5±1.7 & 11.1±0.8 &   & 12.3±2.6 & 10.3±1.0 & 12.9±3.0 & 13.3±0.8 \\
    & \textbf{SWAG}     & 8.3±0.5 & 12.1±2.0 & 11.4±2.5 & 10.2±1.0 &   & 7.9±0.7 & 9.0±1.0 & 11.9±2.3 & 11.2±0.9 &   & 12.3±2.2 & 8.9±0.9 & 11.0±1.9 & 10.5±2.5 \\
    & \textbf{Ensemble} & 6.4±1.6 & 10.3±3.6 & 9.1±3.0 & 6.5±0.6 &   & 4.8±0.4 & 5.1±0.5 & 6.7±2.7 & 7.7±0.8 &   & 5.0±0.6 & 5.6±1.1 & 6.2±1.7 & 7.5±1.8 \\

    \midrule
    \\[-1.5em]
    \cmidrule(lr){3-6}\cmidrule(lr){8-11}\cmidrule(lr){13-16}
    \cmidrule(lr){3-6}\cmidrule(lr){8-11}\cmidrule(lr){13-16}

    % Dataset labels BELOW, with rules that stop exactly under their columns, skipping the ghost column.
    & & \multicolumn{4}{c}{\textbf{LIDC}} & & \multicolumn{4}{c}{\textbf{NPC}} & & \multicolumn{4}{c}{\textbf{Chaksu}} \\
\end{tabular}
}
\caption{Ranking by the entanglement proxy ($\Delta$).}
\label{tab:entanglement_ranking_conf}
\end{subtable}

\caption{Average ranking (1-19) of performance and the entanglement proxy ($\Delta$) when comparing the 19 AU-EU model combinations. The ranks were calculated separately per task and averaged across tasks.}
\label{tab:ranking_tables_conf}
\end{table}

\FloatBarrier

\begin{figure}[htbp]
    \centering
    \includegraphics[width=0.98\textwidth]{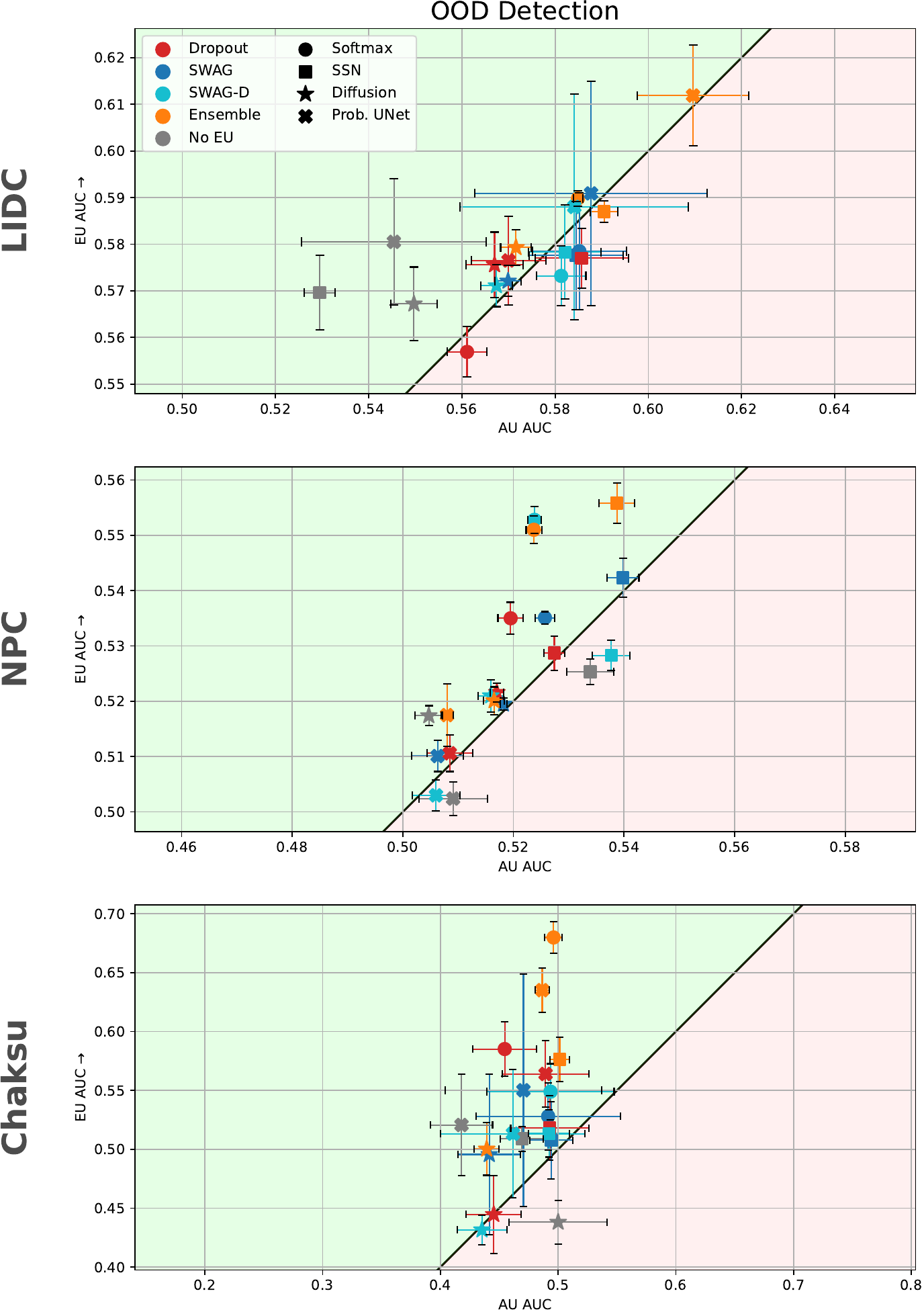}
    \caption{Scatter plots showing the OOD detection performance of model combinations with confidence intervals (otherwise the same as in \cref{fig:scatter_grid_combined}). Each point corresponds to a model combination, with y-values based on the theoretically consistent uncertainty measure and x-values using the inconsistent one.}
    \label{fig:scatter_conf_OODD}
\end{figure}

\begin{figure}[htbp]
    \centering
    \includegraphics[width=0.98\textwidth]{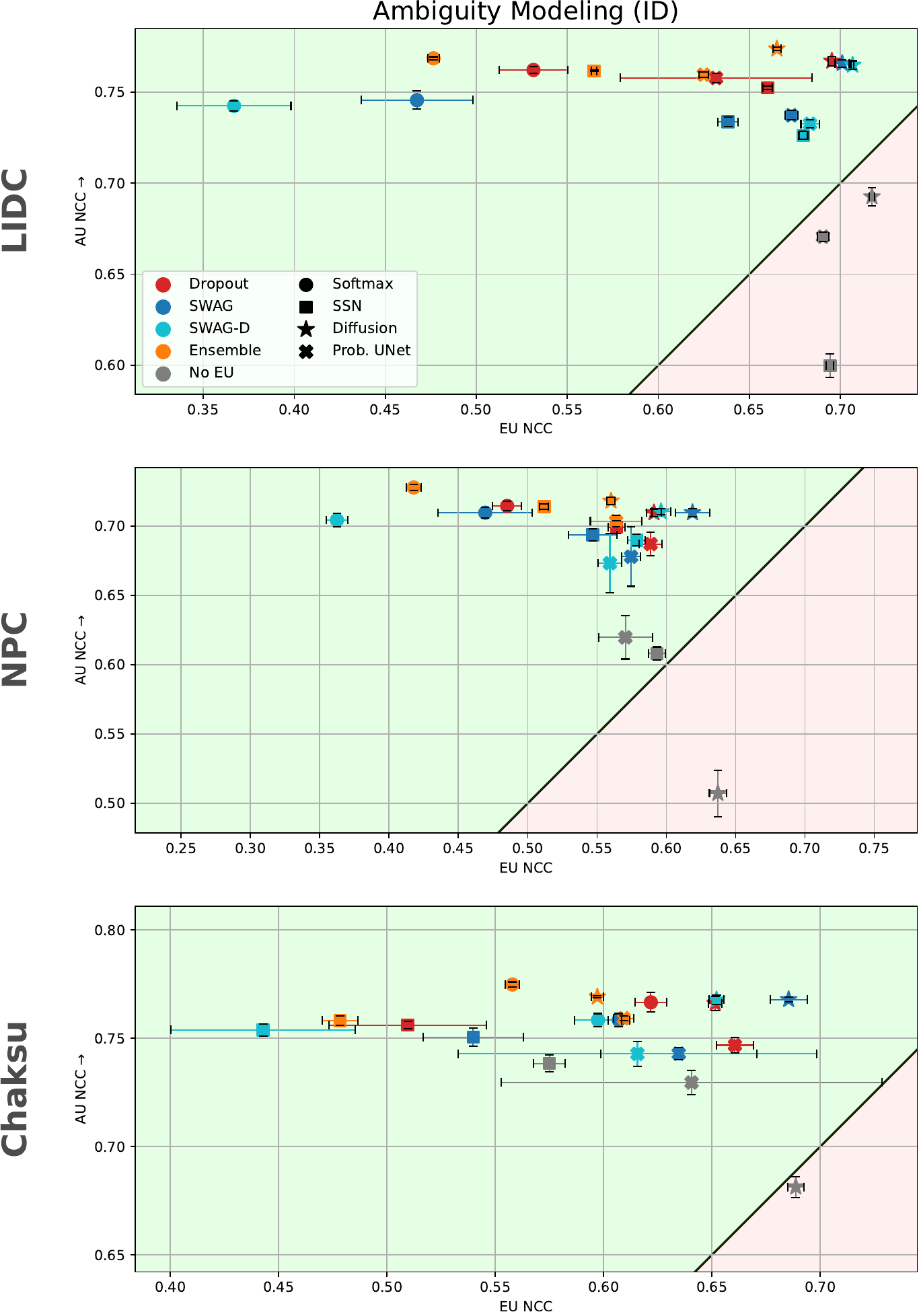}
    \caption{Scatter plots showing the AMB performance of model combinations with confidence intervals (otherwise the same as in \cref{fig:scatter_grid_combined}). Each point corresponds to a model combination, with y-values based on the theoretically consistent uncertainty measure and x-values using the inconsistent one.}
    \label{fig:scatter_conf_AMB}
\end{figure}

\begin{figure}[htbp]
    \centering
    \includegraphics[width=0.98\textwidth]{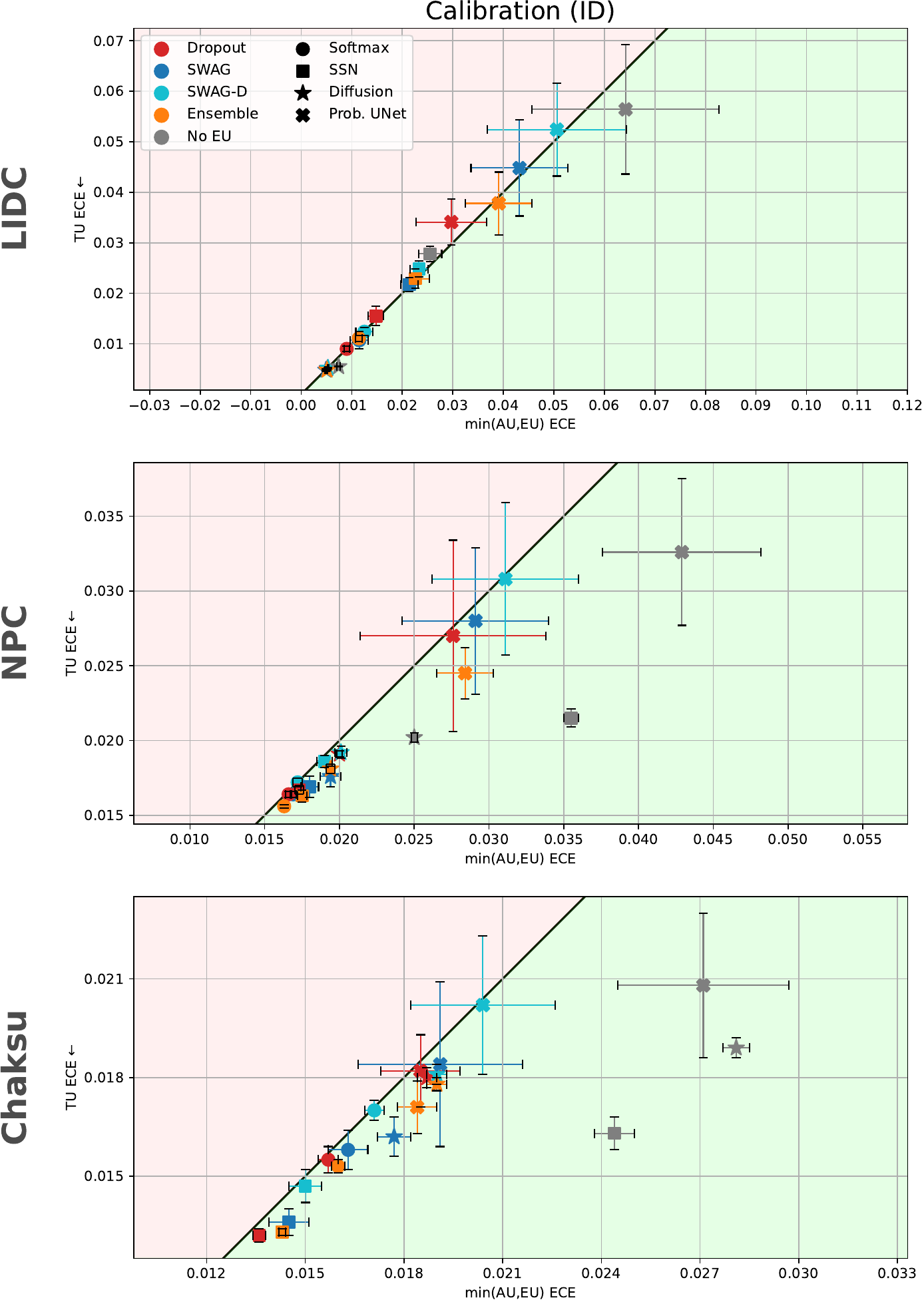}
    \caption{Scatter plots showing the CAL performance of model combinations with confidence intervals (otherwise the same as in \cref{fig:scatter_grid_combined}). Each point corresponds to a model combination, with y-values based on the theoretically consistent uncertainty measure and x-values using the inconsistent one.}
    \label{fig:scatter_conf_CAL}
\end{figure}